\documentclass[letter, 10pt, conference]{ieeeconf}      

\IEEEoverridecommandlockouts                              

\usepackage{graphicx}
\usepackage{amsmath}
\usepackage{amssymb}
\usepackage{subcaption}
\usepackage{booktabs}
\usepackage{hyperref}
\usepackage{makecell}
\let\labelindent\relax
\usepackage{enumitem}

\newcommand{\fig}[1]{Fig.\@~#1}
\newcommand{\tab}[1]{Tab.\@~#1}

\usepackage{array,multirow,graphicx}

\usepackage{tikz}

\usepackage[draft, markup=bfit, deletedmarkup=sout, authormarkup=brackets]{changes}
\colorlet{Changes@Color}{blue}
\definechangesauthor[name={Petr}, color=red]{PS}
\definechangesauthor[name={Matej}, color=orange]{MH}
\definechangesauthor[name={Kuba}, color=magenta]{JR}

\usepackage[firstpageonly=true]{draftwatermark}

\begin{document}
\SetWatermarkAngle{0}
\SetWatermarkColor{black}
\SetWatermarkLightness{0.5}
\SetWatermarkFontSize{9pt}
\SetWatermarkVerCenter{30pt}
\SetWatermarkText{\parbox{30cm}{%
\centering This is the authors' final version of the manuscript published as:\\
\centering Rozlivek, J.; Svarny, P. \& Hoffmann, M. (2023),\\
\centering Perirobot space representation for HRI: measuring and designing collaborative workspace coverage by diverse sensors \\
\centering  in '2023 IEEE/RSJ International Conference on Intelligent Robots and Systems (IROS)', pp. 5958-5965. (C) IEEE \\
\centering \url{https://doi.org/10.1109/IROS55552.2023.10341829}
}}

\title{\LARGE \bf Perirobot space representation for HRI: measuring and designing collaborative workspace coverage by diverse sensors}

\author{Jakub~Rozlivek,
        Petr~Svarny,
        and~Matej~Hoffmann
\thanks{J.R. and M.H. were supported by the Czech Science Foundation (GA CR), project no. 20-24186X. J.R. was additionally supported by the Czech Technical University in Prague, grant no. SGS22/111/OHK3/2T/13. P.S. was supported by the Ministry of Industry and Trade of the Czech Republic (project no. CZ.01.1.02/0.0/0.0/19\_264/0019867).  We thank Tomas Svoboda for his comments on the manuscript.}
\thanks{Jakub Rozlivek, Petr Svarny, and Matej Hoffmann are with the Department of Cybernetics, Faculty of Electrical Engineering, Czech Technical University in Prague. (e-mail: jakub.rozlivek@fel.cvut.cz; matej.hoffmann@fel.cvut.cz).}
}

\maketitle

\begin{abstract}
Two regimes permitting safe physical human-robot interaction, speed and separation monitoring and safety-rated monitored stop, depend on reliable perception of the space surrounding the robot. This can be accomplished by visual sensors (like cameras, RGB-D cameras, LIDARs), proximity sensors, or dedicated devices used in industrial settings like pads that are activated by the presence of the operator. The deployment of a particular solution is often ad hoc and no unified representation of the interaction space or its coverage by the different sensors exists. In this work, we make first steps in this direction by defining the spaces to be monitored, representing all sensor data as information about occupancy and using occupancy-based metrics to calculate how a particular sensor covers the workspace. We demonstrate our approach in two sensor-placement experiments in three static scenes and one experiment in a dynamic scene. The occupancy representation allow the comparison of the effectiveness of various sensor setups. Therefore, this approach can serve as a prototyping tool to establish the sensor setup that provides the most efficient coverage for the given metrics and sensor representations.
\end{abstract}



\IEEEpeerreviewmaketitle

\section{Introduction}\label{sec:intro}

Safety is a fundamental requirement of physical human-robot interaction. While there are multiple approaches to ascertain it~\cite{lasota2017survey}, these were funneled in the industrial context into four collaborative operations by the standard for  industrial robot safety ISO 10218~\cite{ISO10218} with more details in the technical specification for collaborative robots ISO/TS 15066~\cite{ISO/TS15066}.
Two of those operations---safety-rated monitored stop (SRMS) and speed and separation monitoring (SSM)---depend on sensors that monitor the space surrounding the robot. The SRMS specification demands that the system detects the presence of a person within the intended workspace and keeps the robot in a monitored stop state if a person is present. For SSM, it is necessary to maintain at least the protective separation distance between the human and robot at all times.
Unfortunately, in practice, the safety evaluation of the robot application relies on experience and general guidelines, not on a formal approach. 

\begin{figure}[ht]
\begin{subfigure}{0.225\textwidth}
\includegraphics[width=\textwidth]{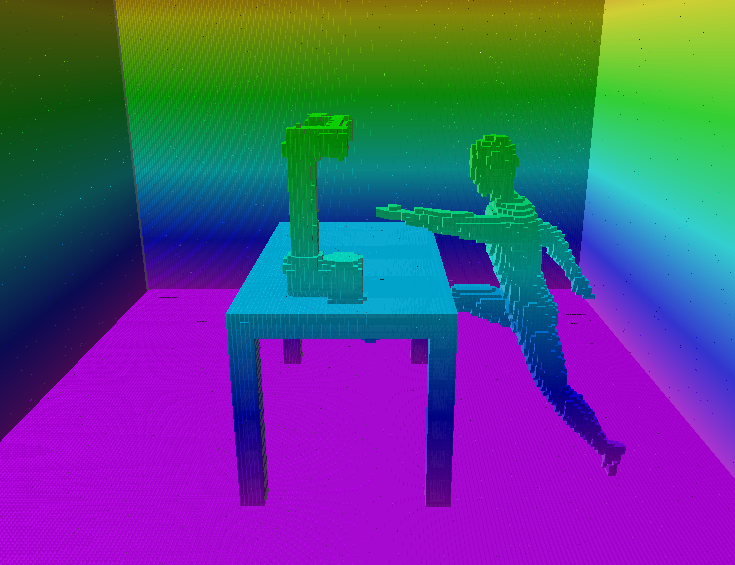}
\caption{\footnotesize Scene 1, human reaching towards the robot.}
\label{fig:case_1}
\end{subfigure}
\hfil
\begin{subfigure}{0.225\textwidth}
\includegraphics[width=\textwidth]{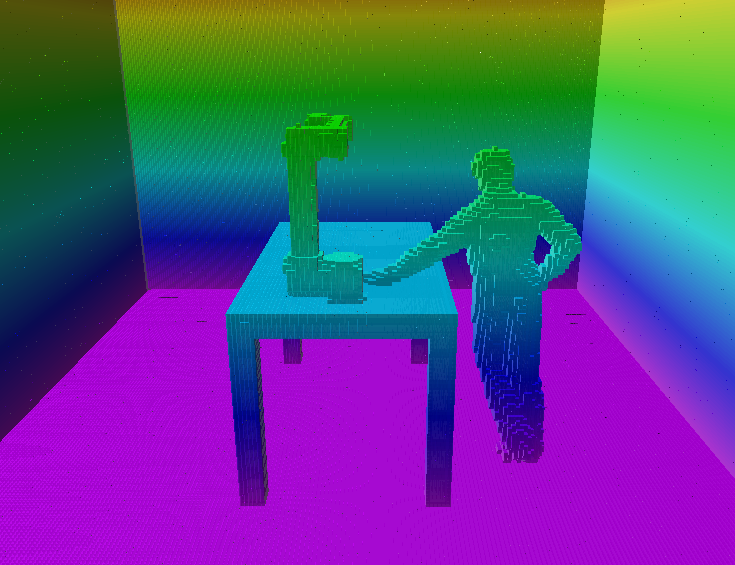} 
\caption{\footnotesize Scene 2, human leaning against the table with the hand.}
\label{fig:case_2}
\end{subfigure}
\begin{subfigure}{0.225\textwidth}
\includegraphics[width=\textwidth]{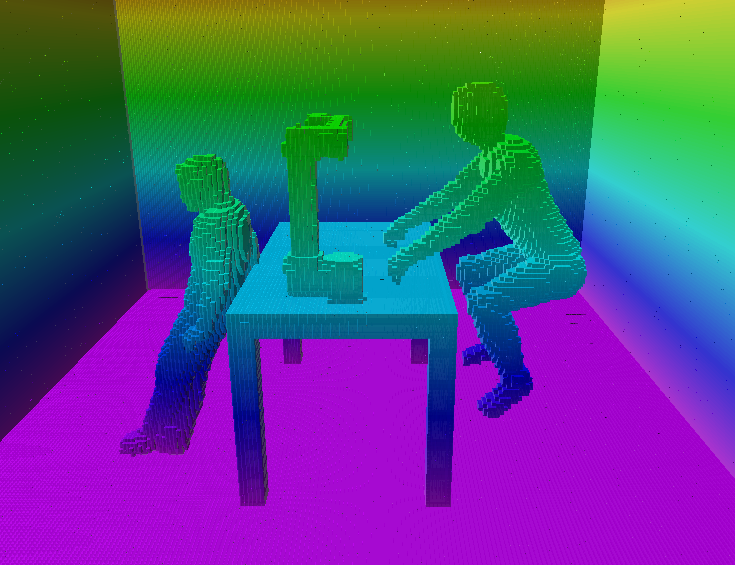}
\begin{minipage}[t][1.1cm][t]{\linewidth}
\caption{\footnotesize Scene 3, human operator without floor contact; another human present.}
\label{fig:case_3}
\end{minipage}
\end{subfigure}
\hfill
\begin{subfigure}{0.225\textwidth}
\includegraphics[width=\textwidth]{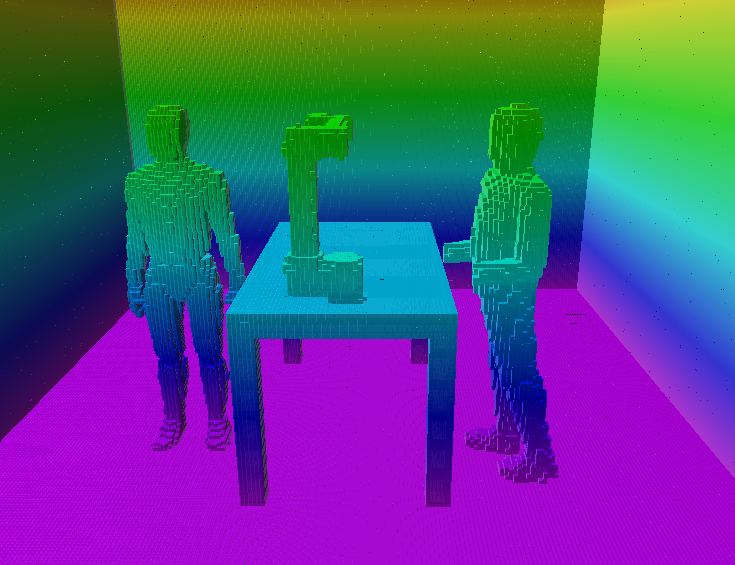}
\begin{minipage}[t][1.1cm][t]{\linewidth}
\caption{\footnotesize Dynamic scene, a human (left) walking around the table and observing the working human.}
\label{fig:case_4}
\end{minipage}
\end{subfigure}
\vspace*{-2mm}
\caption{\small Studied scenes represented as occupancy models. Height is color-coded in the simulation output for better readability.}
\label{fig:use_cases}
\vspace*{-5mm}
\end{figure}

Here we propose to use a unified occupancy-based modeling approach for the space surrounding the robot, i.e., the \textit{perirobot space} (PeRS). This modeling allows comparison of various sensor setups and formal evaluation of their workspace coverage. It can be used to determine the optimal sensor setup given a task, or, if the sensor setup is given, to modify the task or choose its best position and orientation with respect to the sensors.
An illustration of the studied scenarios is in Fig.~\ref{fig:use_cases}.

\noindent\textbf{Contribution.} 
The contributions of this paper are the following:

\begin{itemize}
    \item A unified robot surrounding space representation approach.
    \item Demonstration of the main benefits of this approach: the delimitation of occlusions, explicit definition of sensor data representation, and a clear description of the space monitored by the sensors.
    \item Novel use of established metrics to evaluate coverage of regions of interest.
    \item Application of the approach on an optimal (multi-) sensor coverage task on four different scenes.
    \item A publicly available code repository with the perirobot space implementation\footnote{\url{https://github.com/ctu-vras/perirobot-space}}.
\end{itemize}

\section{Related work}\label{sec:related}

The origin of our research is in safe human-robot interaction (HRI). While much work in safety research has been done on the side of the robot control algorithms, we focus in this article on workspace monitoring. Space that is not monitored is unknown to the robot, and any activity (or lack thereof) cannot affect the robot's behavior. Four fields of research are relevant to this problem.

First, sensor coverage approaches are relevant (see, for example, \cite{wang2011coverage}, \cite{elhabyan2019coverage}). Nevertheless, they often aim at coverage from a theoretical perspective and do not combine multiple types of sensors. An exception in this respect is the recently published work by Oščádal et al.~\cite{Oscadal2023}. They present an approach where they determine the importance of individual voxels in the shared human-robot workspace and arrange cameras to provide the highest coverage of the monitored space. However, they do not take into account occlusions caused by the human or other objects in the scene.

The second research field, occlusion mitigation or gaze control, as in \cite{he2022visibility}, \cite{ibrahim2022whole} or \cite{finean2021should}, is also related as it shares the aim to monitor a region of space efficiently. However, this region, as opposed to coverage approaches, is small (usually merely a target point). For safe interaction, all the relevant robot surrounding space needs to be considered.

Sensor fusion approaches, the third relevant research area, do not address coverage, but focus on the fusion of various sensor inputs (see \cite{alatise2020review}). Still, safety-related works such as \cite{Magnanimo2016, rashid2020local} leverage sensor fusion to ascertain sufficient workspace coverage. An interesting addition in this respect is \cite{suvsanj2020effective} which evaluates two types of sensors (albeit not used together in the same setup) for 2D and 3D coverage.

The fourth field of research is SSM-motivated workspace monitoring, where methods of implementing efficient sensing and distance measurement algorithms are studied (e.g., Marvel~\cite{Marvel2013, marvel2017implementing}). Specific aspects such as investigations of the nature of the relative distances between human and robot keypoints are studied (see \cite{fabrizio2016real} for a comparison of approaches and their approach based on the depth field). Studies of the obstacle representation are also relevant: from the kinetostatic danger field to sphere swept lines-based bounding volumes \cite{Magnanimo2016, Lacevic2010, Zanchettin2016, Polverini2017, byner2019dynamic, scalera2021optimal}. These approaches, however, focus on robot control and not monitoring itself.

Finally, let us mention that the majority of the presented papers investigate camera-based monitoring (see also \cite{halme2018review}). However, there are other alternatives for safety-related monitoring, e.g., time-of-flight sensors~\cite{kumar2019speed} or proximity sensing~\cite{qi2022safe}. Especially proximity sensing presents a promising and recent avenue of research and focuses on robot-mounted sensors that can detect obstacles up to 0.5 m from the robot surface; see \cite{Navarro2021} for a thorough review.

\section{Proposed approach}\label{sec:proposed}
Our interest in the space surrounding the robot is motivated by safe HRI. The inspiration comes from humans and their peripersonal space---a dynamically established protective safety margin around the whole body, drawing on visuo-tactile interactions (see \cite{Roncone2016} for more details and an implementation on a humanoid robot). However, perirobot space aims to be a more general approach that is not limited by the human-likeness of peripersonal space (e.g., sensors are not limited only to the robot's body). We first discuss the region of interest for the proposed approach. Then we present the modeling and representation of sensors and the notion of perirobot space itself. Finally, we present the evaluation metrics and the modeling approach.

\subsection{Region of interest}\label{subsec:perirobot}
We study a given region of interest, the volume where HRI can occur. 
The surrounding space of the robot can be defined following the definitions from \cite[Chapter~3]{siciliano2008springer}.  For example, we could distinguish either the robot's workspace (i.e., positions reachable by the end-effector) or the robot's envelope (i.e., the total volume of space occupied by the robot during these positions). There are even more elaborate approaches to capture the space surrounding the robot (see also Sec.~\ref{sec:related}).

\begin{figure}[t]
\centering
\begin{subfigure}{0.47\textwidth}
\centering
\includegraphics[width=0.9\textwidth]{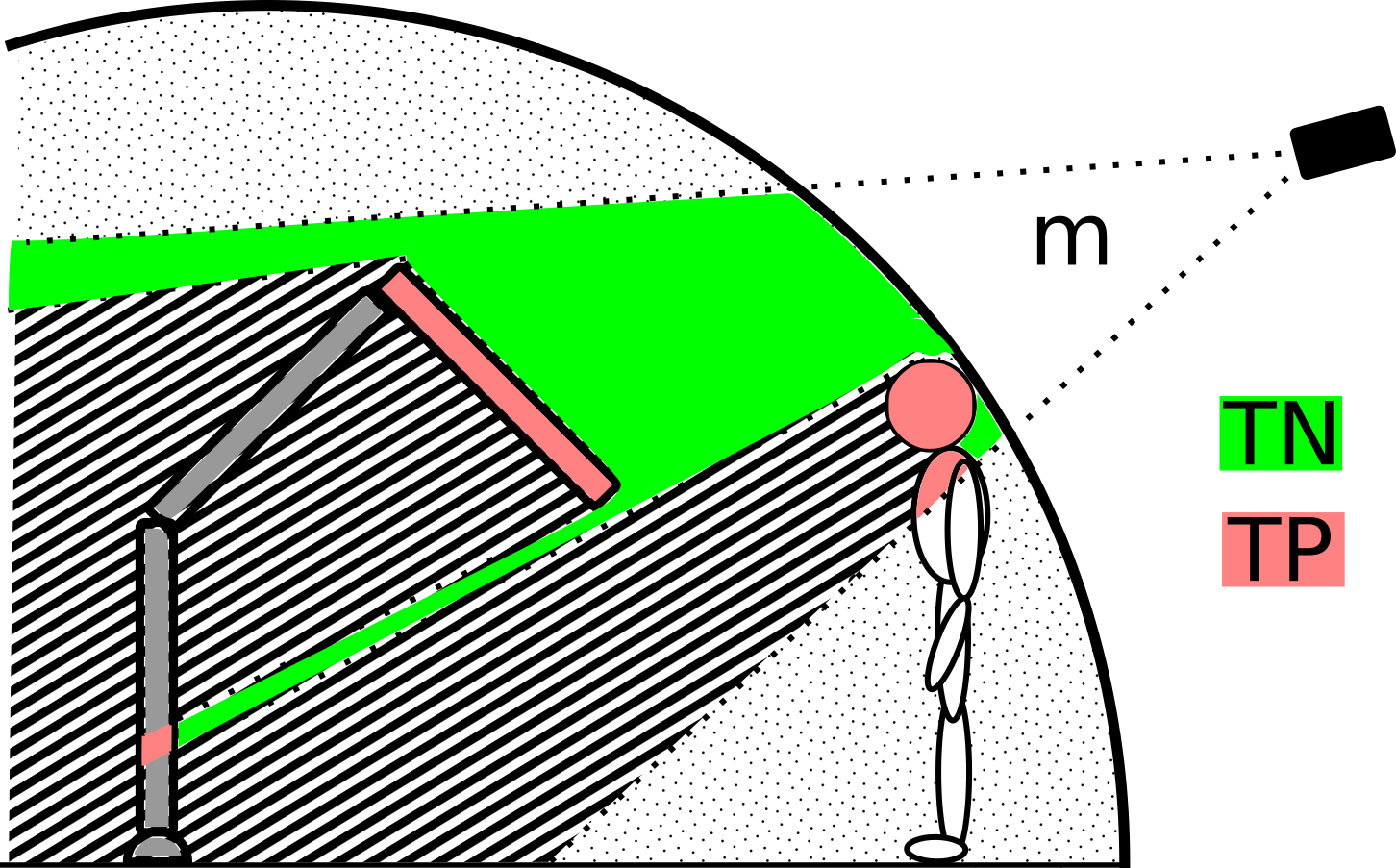} 
\caption{\footnotesize Schema for the semi-spherical robot-centered region of interest.}
\label{fig:pers_diagram_robot}
\end{subfigure}

\begin{subfigure}{0.47\textwidth}
\centering
\includegraphics[width=0.9\textwidth]{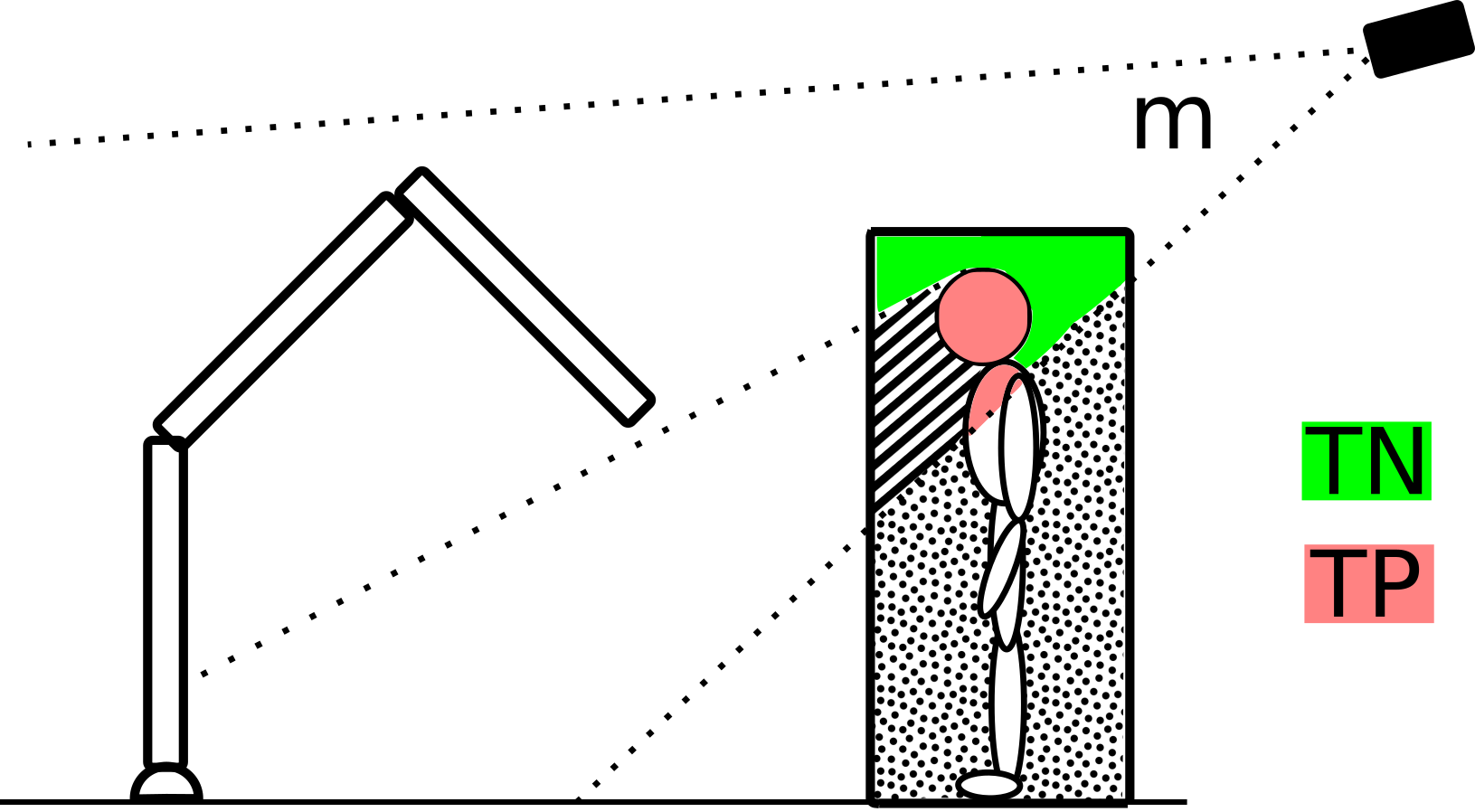} 
\caption{\footnotesize Schema for the human bounding box region of interest.}
\label{fig:pers_diagram_human}
\end{subfigure}

\caption{\small Two regions of interest with a single sensor---a camera---and its perception space $m$.
We distinguish four types of the space in the region of interest: occupied (red), free (green), unknown monitored free space (striped), unknown monitored occupied space (grey) and not monitored free space in the region of interest (dotted). 
Note that the occupied space is merely for illustration purposes this large. 
In practice, only the surfaces of the captured objects are registered.}
\label{fig:pers_diagram}
\vspace*{-2mm}  
\end{figure}

In our approach, we study one robot-centered and one human-centered region. The robot-centered region is defined simply as the (semi-)sphere centered at the robot whose radius is given by the robot's end-effector's maximal reach, see \fig{\ref{fig:pers_diagram_robot}}. This definition of the `robot' space signifies that for safe interaction it is necessary to monitor the robot's full reach.
We designate as the `human' space the bounding box enclosing the human that is next to the robot, see space $h$ in \fig{\ref{fig:pers_diagram_human}}. 
By this representation, it is meant that this space needs to be efficiently monitored to ascertain safe interaction.

\subsection{Surrounding space monitoring}
Monitoring the surrounding space together with the robot's properties determine how the robot would be controlled. For example, a sensor that should trigger the robot's stop at full speed in order to prevent a collision needs to be placed so that it gives the robot sufficient time to brake. Therefore, sensor information should also be part of the representation of the surrounding space of the robot. However, sensor monitoring also introduces a new challenge---occlusions and especially sensor data representation.

The data from the sensors monitoring the robot surrounding space are always represented in some manner. 
This representation is a deliberate choice and is not defined only by the collected raw sensor data itself. 
For example, a line laser sensor can be both a switch (i.e., something passed in front of the sensor) or it can serve as a so-called profiler (i.e., measure the profile of the object under the laser).

Additionally, the representation of the sensor reading is specific to the robot, the task, and the available sensors. 
The choice of representation can lead to different control decisions for the robot. 
For example, the collision classification~\cite{Haddadin2017} for a pressure sensor on the robot can have at least two representations.
A collision can be interpreted as a stop signal or as an impulse to move away.

\subsection{Sensor modeling}\label{subsec:sensors}
Our approach is to model the representation of the sensor readings as information about space occupancy. 
While safety-related research focuses on detecting the human, we aim to represent the space surrounding the robot as a whole as it is perceived by the sensors. 
Therefore our approach represents any relevant sensor data as occupied space, be it a human, robot or an object. 
We can distinguish three main ways of sensor representation:
\begin{itemize}
\item{\textbf{Naive representation.}} Pure occupancy information provided from the sensor.
\item{\textbf{Volume-based.}} The sensor information is represented as occupancy of a predetermined volume (e.g., pressing a floor pad leads to the assumption that the whole space above the pad is occupied).
\item{\textbf{Feature-based.}} Sensor data are used to determine features (often human keypoints) and their locations. 
This representation can be further extended by determining a bounding box around the human, the feature neighborhoods (e.g., surrounding spheres), or volumes between features (e.g., cylinders connecting human keypoints) which are all considered as occupied space.
\end{itemize}


\begin{figure}[h]
\begin{subfigure}{0.23\textwidth}
\includegraphics[width=\textwidth]{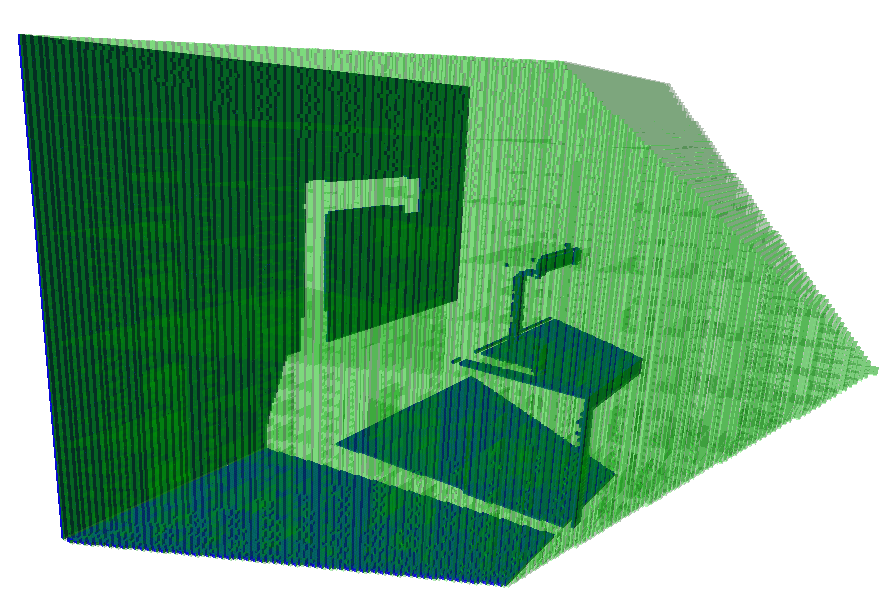} 
\begin{minipage}[t][0.5cm][t]{\linewidth}
\caption{\footnotesize RGB-D camera (without noise)}
\label{fig:camera_occupancy}
\end{minipage}
\end{subfigure}
\hfill
\begin{subfigure}{0.23\textwidth}
\includegraphics[width=\textwidth]{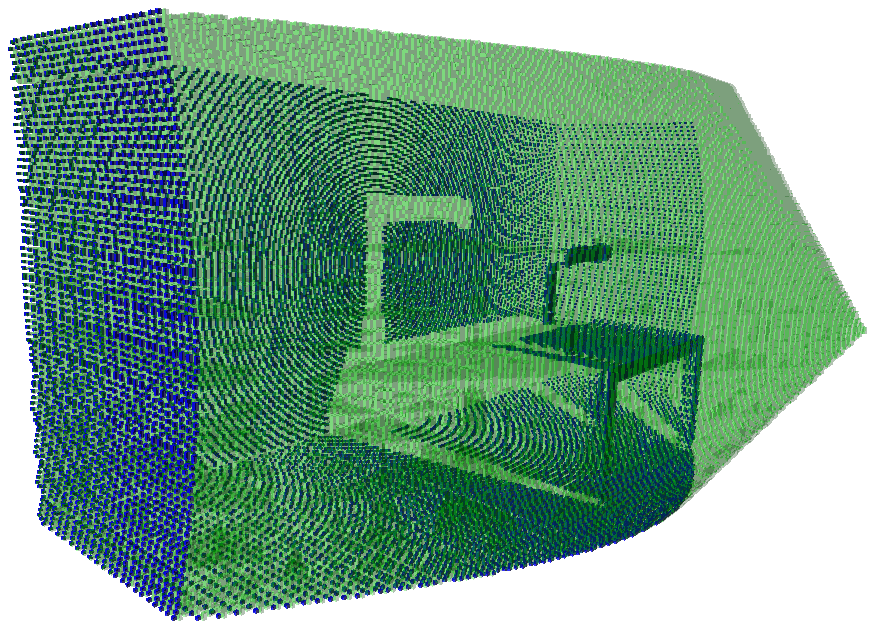} 
\begin{minipage}[t][0.5cm][t]{\linewidth}
\caption{\footnotesize LIDAR sensor (without noise)}
\label{fig:lidar_occupancy}
\end{minipage}
\end{subfigure}
\begin{subfigure}{0.5\textwidth}
\centering
\includegraphics[width=0.5\textwidth]{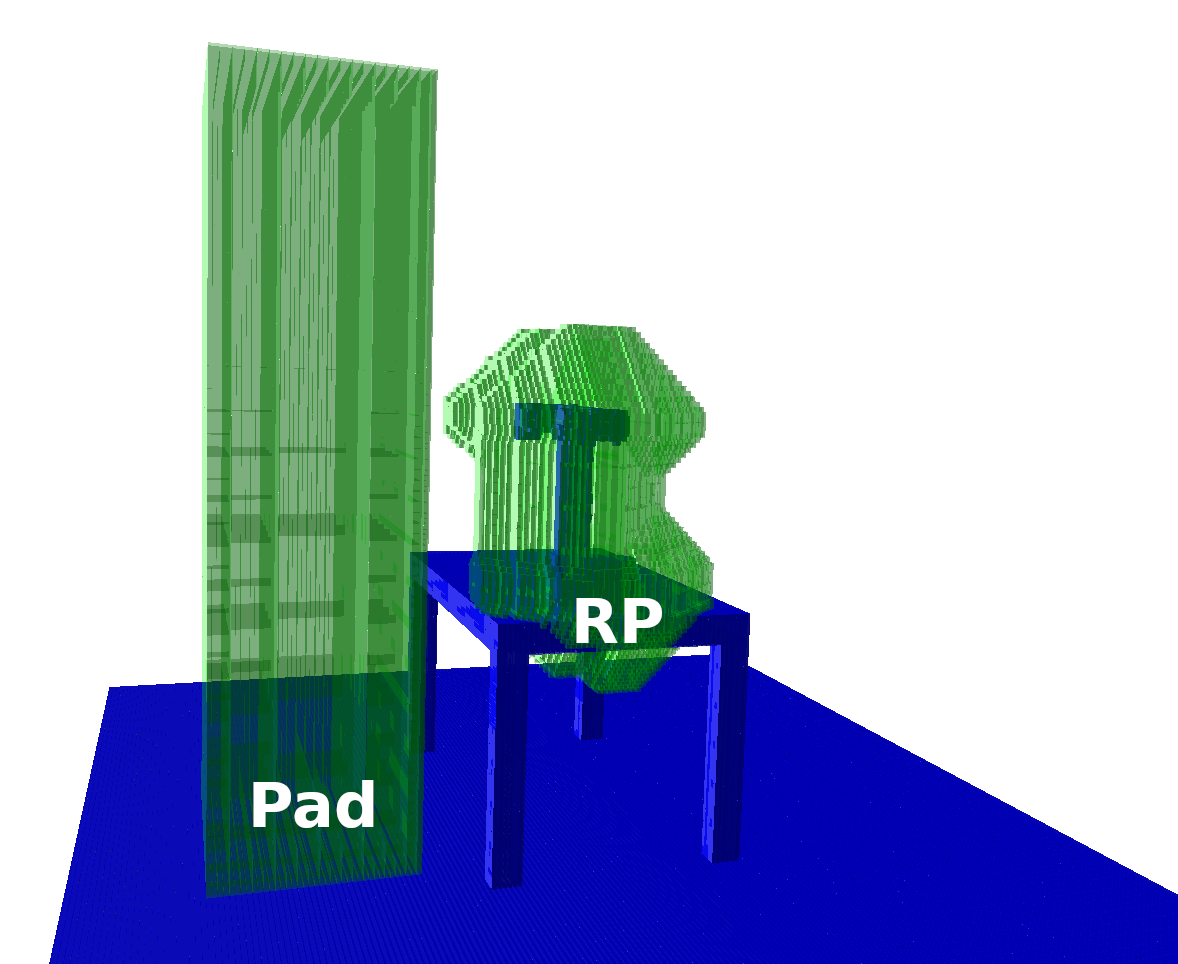} 
\begin{minipage}[t][0.5cm][t]{\linewidth}
\caption{\footnotesize Pressure pad (Pad) and robot proximity cover (RP).}
\label{fig:sensor_occupancy_model}
\end{minipage}
\end{subfigure}
\caption{\small Visualization of occupancy based representations. The space monitored by the sensors is in green, while obstacles are in blue.}
\label{fig:sensor_viz}
\vspace*{-3mm}
\end{figure}

We analyze five sensors in this study, four of which are shown in Fig.~\ref{fig:sensor_viz}.
Our assumption is that thanks to the simple representation, any relevant sensor could be represented in a similar manner. Namely, each sensor fits roughly into one of the three types:

\begin{itemize}
\item{\textbf{Area.}} Sensors activated by events in a given area in the workspace. For example the pressure pad is activated by stepping on a particular area.
\item{\textbf{Range.}} Sensors monitoring the depth of the space around.
\item{\textbf{Zone.}} Sensors safeguarding a given zone but not necessarily monitoring the zone itself, e.g., a gate monitoring the entrance to the zone. 
\end{itemize}

All of the analyzed sensors are defined by their 6D poses and additional defining parameters. The sensors and their defining parameters are the following:
\begin{itemize}
\item{\textbf{RGB camera.}} Pinhole camera model, simulated as ray-casting, without the depth information provided. Parameters: field of view, resolution.
\item{\textbf{RGB-D camera.}} Pinhole camera model, simulated as ray-casting, see \fig{\ref{fig:camera_occupancy}}. Parameters: field of view, resolution. 
\item{\textbf{LIDAR sensor.}} Arc of rays, see \fig{\ref{fig:lidar_occupancy}}. Parameters: field of view, range.
\item{\textbf{Pressure pads.}} Defined by their area and active if there are any contact points, i.e., occupied voxels right above the pad, designates all volume above the pad as occupied, see `P' in \fig{\ref{fig:sensor_occupancy_model}}. Parameters: dimensions. 
\item{\textbf{Robot proximity cover.}} Inflation of the robot model volume (inflation value), see `RP' in \fig{\ref{fig:sensor_occupancy_model}}.
\end{itemize}

The combinations of sensors and representations used in this paper are listed in Tab.~\ref{tab:sensor_represent}.

\begin{table}[htb]
\centering
\resizebox{250pt}{!}{%
\begin{tabular}{r|c|ccc}
     \multirow{2}{*}{Sensor} & \multirow{2}{*}{Type}   & \multicolumn{3}{c}{Representation} \\
 &  & Naive         & Volume & Feature        \\[1mm] \hline
RGB Camera & Zone &  &        & x \\
RGB-D Camera & Range & x &          & x \\
LIDAR & Range  & x             &         &          \\
Pressure pad & Area  &              & x        &          \\
 Proximity & Range    &   x            &          &         
\end{tabular}
}
\caption{\small Representations and types of sensors.}
\label{tab:sensor_represent}
\vspace*{-2mm}
\end{table}


\subsection{Perirobot space}

The \textit{perirobot space} (PeRS) arises from the combination of a region of interest and occupancy-represented sensor data. 
A simplified representation of PeRS is in \fig{\ref{fig:pers_diagram}}. 
This schema shows the monitored space $m$ of the camera and two regions of interest, the robot and human space. 
Notice, however, that the robot itself changes what space is monitored by the camera due to occlusions, and thus the capability to represent its surroundings. 






\subsection{Coverage metrics} \label{subsec:metrics}
We formulate the space coverage problem as a multi-class classification problem with three classes: positive (occupied space), negative (free space), and unknown (not monitored). The last class appears only in the labels as output from our framework. As the data are highly imbalanced (much more free space than the occupied one), we chose two established metrics for multi-class imbalanced classification tasks to evaluate the efficiency of the established PeRS in a specific region of interest. 

\begin{figure}[t]
\centering
\includegraphics[width=0.4\textwidth]{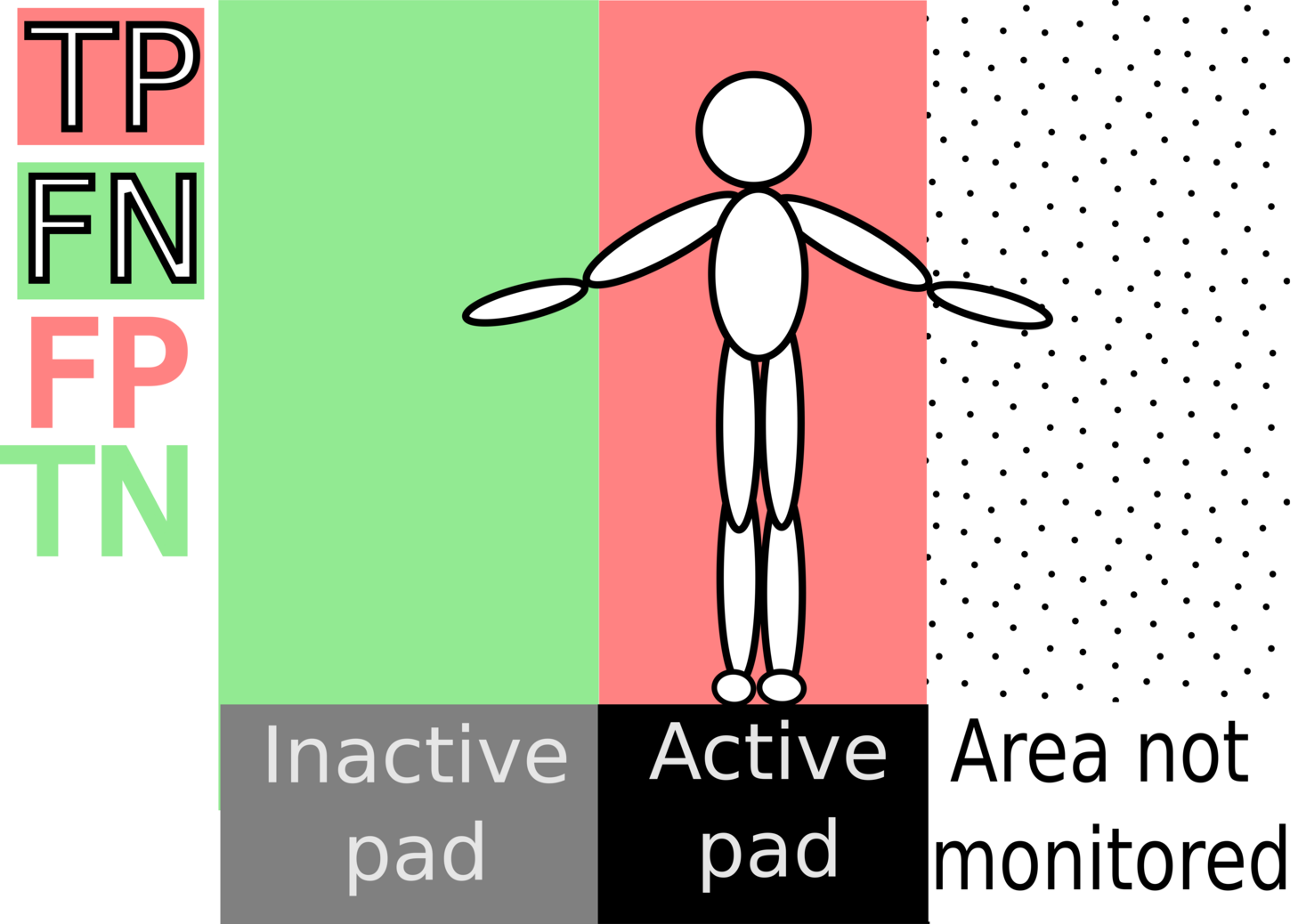} 
\caption{\small Example of the coverage metric representation for a monitored volume with two pads. The active pad designates all the volume above it as occupied. Only the voxels occupied by the person are truly occupied (True Positive, TP, white on red), the voxels not occupied by the person are falsely marked as occupied (False Positive, FP, red). The volume above the inactive pad is considered empty. The majority of the voxels are truly empty (True Negative, TN, green), but the human arm voxels are falsely considered as empty space (False Negative, FN, white on green). Last, there is also space not monitored by the pads containing empty space (dotted) and occupied space (white on dots).}
\label{fig:FTPN_diagram}
\vspace*{-2mm}  
\end{figure}

\textbf{F-score metric.} The first metric is F-score:
\begin{equation}
\label{eqn:f1score}
    F_1 = \frac{2\mathrm{TP}}{2\mathrm{TP}+\mathrm{FP}+\mathrm{FN}+\mathrm{UF} +\mathrm{UO}}
\end{equation}
where UF and UO represent the not monitored free and occupied space, respectively. TP, FP, FN are defined in \fig{\ref{fig:FTPN_diagram}}. This metric does not take into account truly empty voxels. 

\textbf{Cohen's Kappa $\kappa$ metric.}
Since both the truly occupied and truly empty voxels are important in our case, we chose another metric that takes into account both of them. 
Cohen's Kappa is given by the ratio of the relative observed agreement $p_o$ and $p_e$, the hypothetical probability of chance agreement which is defined in multi-class version as:
\begin{equation}
\label{eqn:kappa}
    \kappa =  \frac{s(TP+TN) - \sum_kp_kt_k}{s^2-\sum_kp_kt_k}
\end{equation}
where $s$ is the total number of voxels, $p_k$ is the number of voxels where class $k$ is predicted and $t_k$ is the true number of voxels of class $k$.

\subsection{OctoMap modeling}\label{sec:experiment}
The core notion of our approach is occupancy in the surrounding space. All sensors are represented as providing information about the occupancy of the space they perceive. 
Specifically, we model the scene and, thereafter, all sensors using OctoMap~\cite{hornung2013octomap} and octrees.

OctoMap and the occupancy representation also allow us to model the limitations of various approaches. 
For example, camera occlusions are considered because ray-casting from the simulated sensor is blocked by occupied space. 
Ray-casting also simulates the limited field of view of the cameras.
Additionally, we introduced noise for the range-based sensors (LIDAR, RGB-D camera, and proximity cover) added to the measured distance.

\begin{figure}[h]
\begin{subfigure}{0.1\textwidth}
\includegraphics[width=\textwidth]{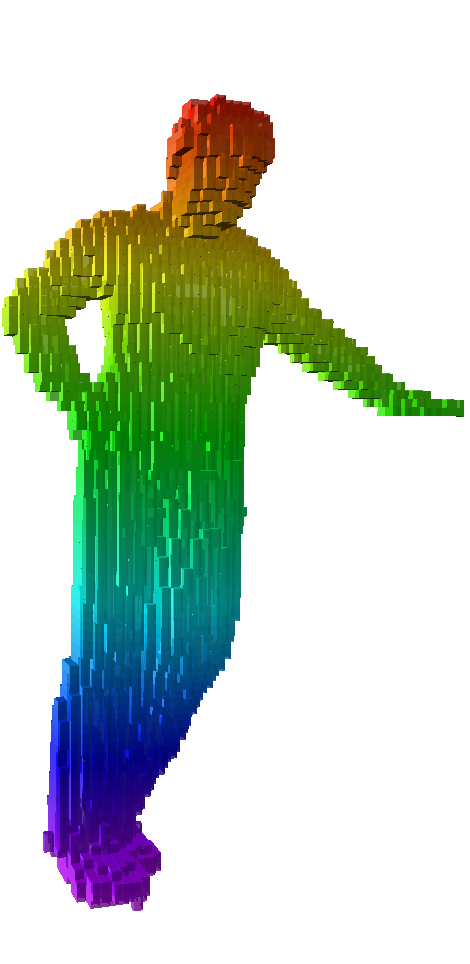}
\begin{minipage}[t][1.2cm][t]{\linewidth}
\caption{\footnotesize \\Ground truth.}
\label{fig:keypoints_human}
\end{minipage}
\end{subfigure}
\hfil
\begin{subfigure}{0.1\textwidth}
\includegraphics[width=\textwidth]{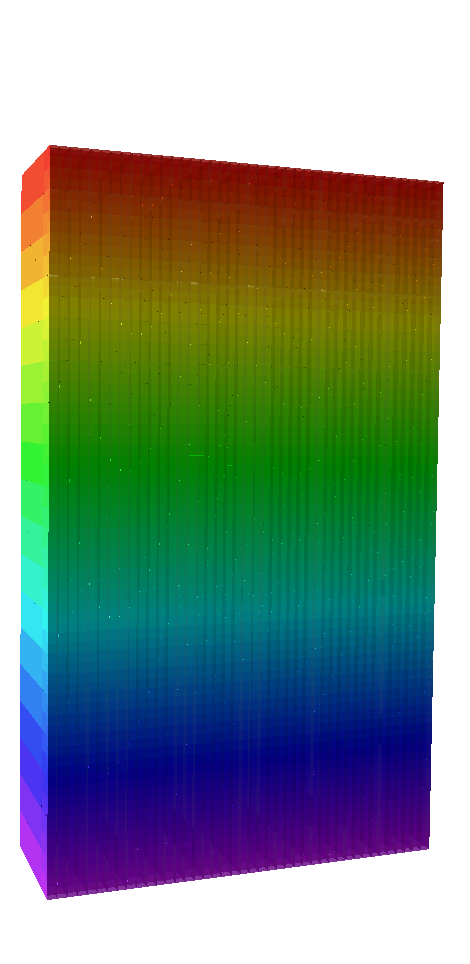} 
\begin{minipage}[t][1.2cm][t]{\linewidth}
\caption{\footnotesize \\Bounding box.}
\label{fig:keypoints_bb}
\end{minipage}
\end{subfigure}
\hfil
\begin{subfigure}{0.1\textwidth}
\includegraphics[width=\textwidth]{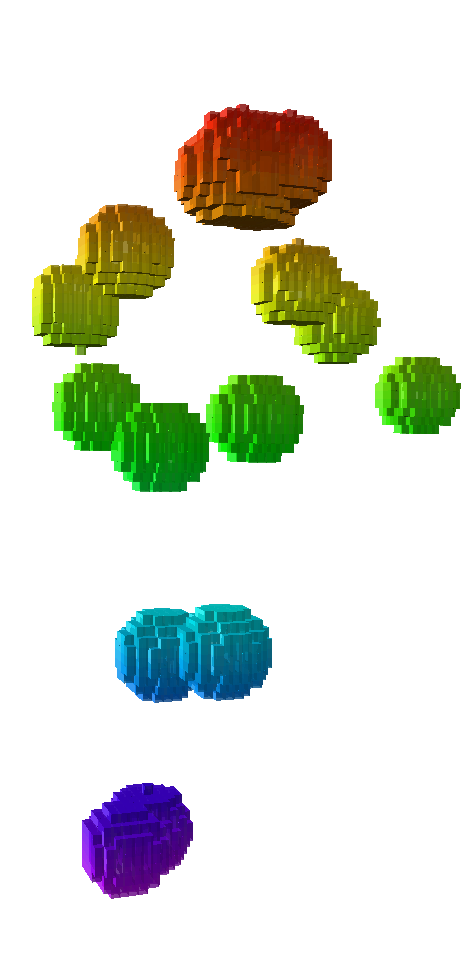}
\begin{minipage}[t][1.2cm][t]{\linewidth}
\caption{\footnotesize \\Spheres.}
\label{fig:keypoints_sp}
\end{minipage}
\end{subfigure}
\hfil
\begin{subfigure}{0.1\textwidth}
\includegraphics[width=\textwidth]{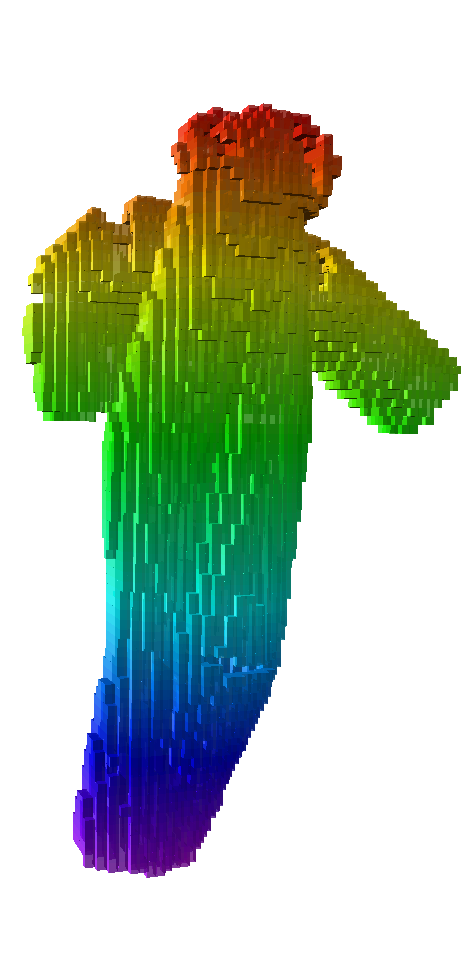}
\begin{minipage}[t][1.2cm][t]{\linewidth}
\caption{\footnotesize \\Cylinder connections.}
\label{fig:keypoints_cyl}
\end{minipage}
\end{subfigure}
\vspace*{-3mm}
\caption{\small Ground truth and human models for the used keypoint representations. Height is color-coded in the simulation output.}
\label{fig:keypoints}
\vspace*{-0mm}
\end{figure}

As mentioned earlier, feature-based sensors create a representation of the detected human keypoints. We distinguish three models of the detected human keypoints based on the usual practice in safe HRI. Namely, a bounding box encompassing all the keypoints, the detected keypoints enclosed in spheres or cylindrical connections between the keypoints. These are shown in \fig{\ref{fig:keypoints}}.

Our modeling approach consists of the following steps:
\begin{enumerate}
    \item Create a model of the scene and determine the region of interest.
    \item Simulate sensor data.
    \item Calculate the sensor-generated occupancy.
    \item Calculate the metrics. (see Sec.~\ref{subsec:metrics})
\end{enumerate}

In our initial approach, we model a single manipulator placed on a table and a human in various poses next to it. 
We investigate two regions of interest: robot space and human space. 
While these could be determined in a task-dependent fashion, we chose regions that can be defined in a straightforward way for demonstration purposes.

\section{Results}\label{sec:results}
To demonstrate our approach, we present three experiments evaluated in a series of HRI scenes (shown in \fig{\ref{fig:use_cases}}).
Although the presented scenes focus on HRI, the problem can be understood in general, as the robot's workspace can contain unforeseen objects that could harm the robot or be damaged by the robot. 
In the first experiment, we simulate an RGB-D camera in three static scenes. The same static scenes are used in the second experiment as well, this time we simulate three different sensors---a pressure pad, a robot proximity cover, and an RGB-D camera. The third experiment is a simulation of a LIDAR sensor in the dynamic scene. The parameters of the sensors are in \tab{\ref{tab:params}}.
\begin{table}[htb]
\centering
\resizebox{250pt}{!}{%
\begin{tabular}{c|c|c|c}
Exp. (Scenes) & Sensor (\# of poses)   & Sensor parameters   & Value\\[1mm] \hline \hline 
&&&\\[-2mm]
\multirowcell{3}{1\\(1,2,3)} & \multirowcell{3}{RGB(-D) camera \\ (172 $\times$ 5 ori)} & FoV  (hor $\times$ ver) [$^{\circ}$] &  87 $\times$ 58  \\
&  & Res. (hor $\times$ ver) [px]  &  1280 $\times$ 720 \\
&  & Range (RGB-D only) [m]  &  0.6 - 6 \\[1mm] \hline
&&&\\[-2mm]
\multirowcell{5}{2\\(1,2,3)} & \multirow{1}{*}{Pressure pad (24)} & Dimensions [m] &  1.0 $\times$ 0.75 \\[1mm] \cline{2-4}
&&&\\[-2mm]
 & Robot proximity cover & Inflation [m]  & 0.1/0.2/0.3 \\[1mm] \cline{2-4}
&&&\\[-2mm]
& \multirow{3}{*}{RGB-D camera (9)} & FoV  (hor $\times$ ver) [$^{\circ}$] &  87 $\times$ 58  \\
&  & Res. (hor $\times$ ver) [px]  &  1280 $\times$ 720 \\
&  & Range [m]  &  0.3 - 3 \\ \hline
&&&\\[-2mm]
\multirowcell{3}{3\\(Dynamic)} & \multirow{3}{*}{LIDAR sensor (172)} & FoV (hor $\times$ ver) [$^{\circ}$] &  360 $\times$ 45  \\
& & Range [m]    &  0.5 - 20  \\ 
& & Ang. res. (hor / ver) [$^\circ$]    &  0.7 / 0.7  \\ \hline
\end{tabular}
}
\caption{\small Experiment parameters.}
\label{tab:params}
\vspace*{-5mm}
\end{table}

\subsection{Different data interpretation experiment (Exp 1)}
In this experiment, we compare the performance of the RGB-D camera observing the workspace based on different interpretations of the measured data. 
Those interpretations are: 1) 2D keypoint detection only---if a human is detected, the defined space around the robot is marked as occupied; 2) Raw 3D point cloud; 3) Raw 3D point cloud with added points from the keypoints surroundings (spheres); 4) Raw 3D point cloud with added points from keypoints cylinder connections; 5) Raw 3D point cloud with added points from the keypoints bounding box. The keypoint representations are shown in \fig{\ref{fig:keypoints}}. 

\begin{figure}[htbp]
    \centering
    \includegraphics[trim={7cm 3cm 0 3cm},clip,width=0.48\textwidth]{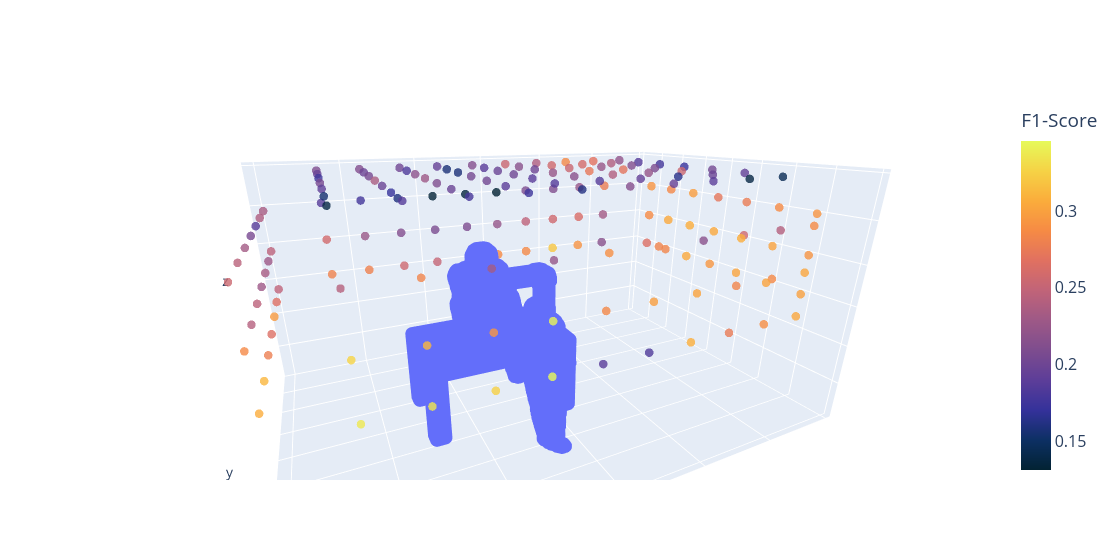}
    \includegraphics[trim={7cm 3cm 0 3cm},clip,width=0.48\textwidth]{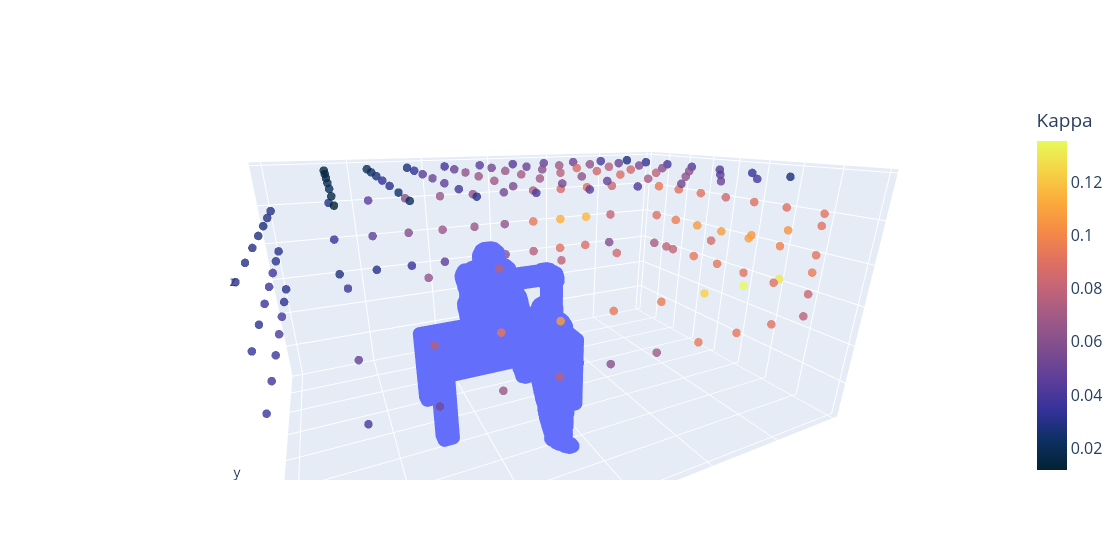}
    \caption{\small RGB-D camera placement heatmap for `robot' space in Scene 3; $F_1$ (top) and $\kappa$ (bottom) values.}
    \label{fig:exp1_heatmap}
    \vspace*{-4mm}
\end{figure}

We placed the RGB-D camera in 172 different positions covering all four walls and ceiling (see \fig{\ref{fig:exp1_heatmap}} for all positions) and rotated the camera to five different orientations in each position, resulting in 860 measurement poses. We computed the $F_1$ and $\kappa$ scores for each measurement pose for the `robot' and `human' spaces and evaluated the distribution of the metrics for each scene to evaluate the performance of each interpretation. 

\begin{figure}[htbp]
    \centering
    \includegraphics[width=0.45\textwidth]{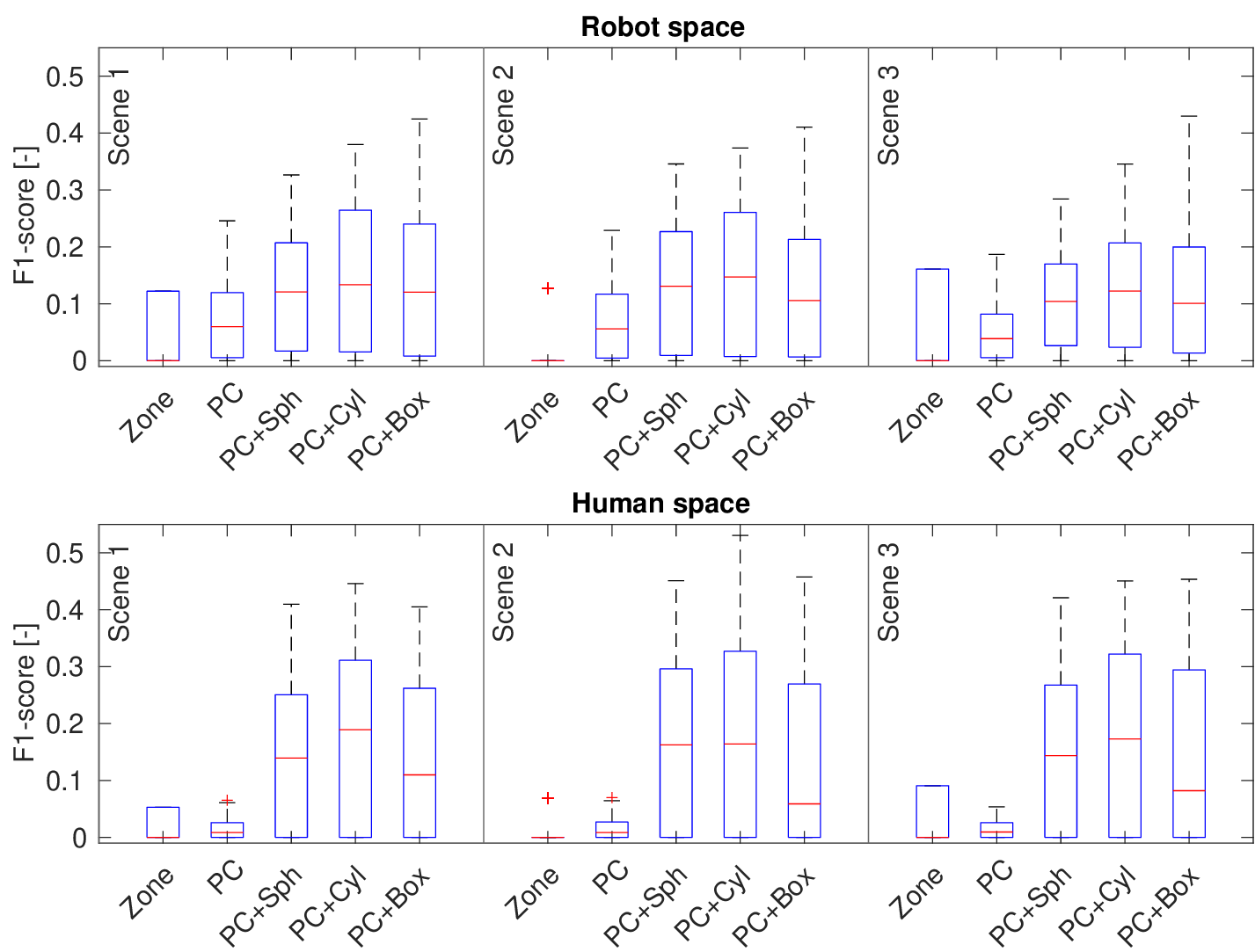}
    \includegraphics[width=0.45\textwidth]{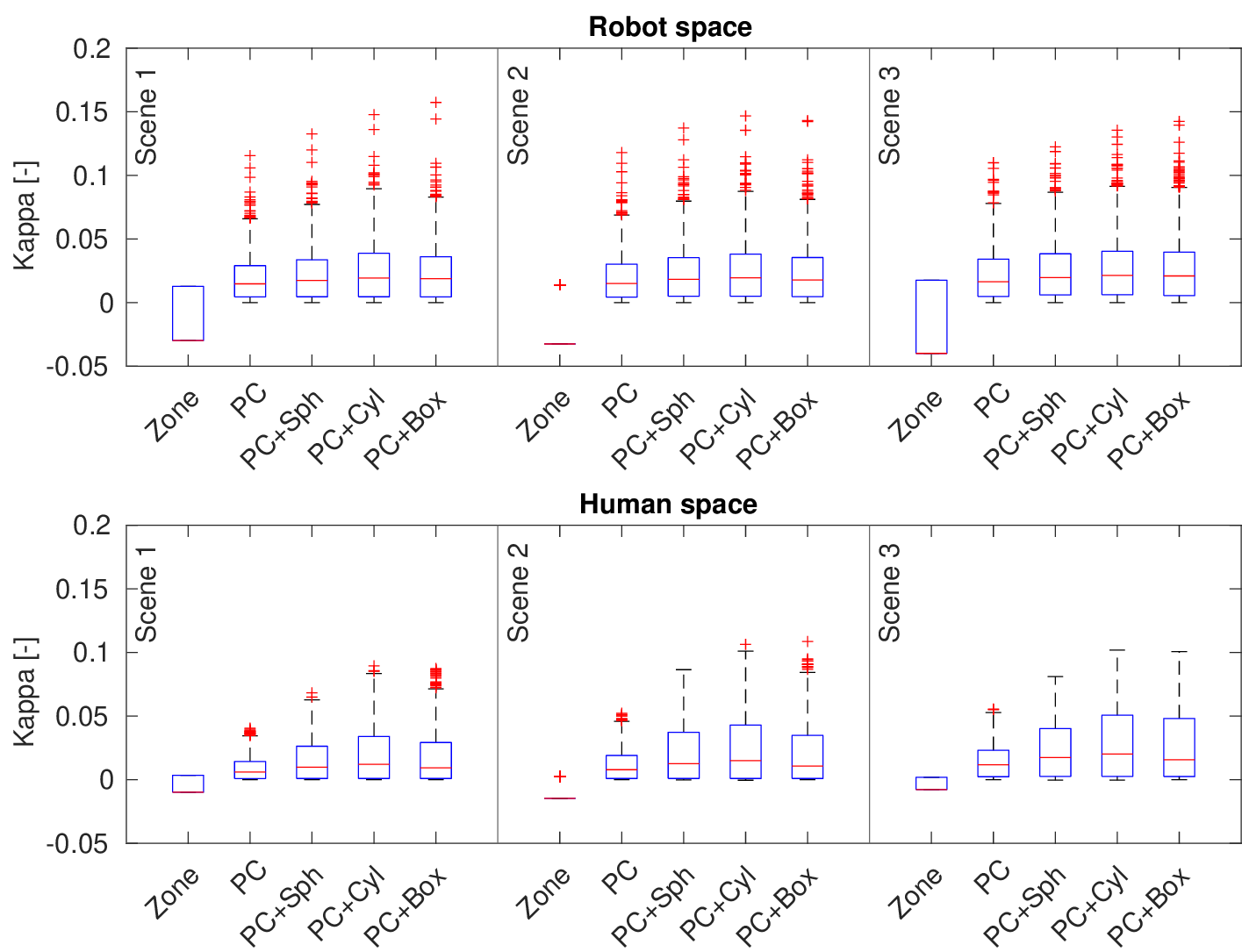}
    \caption{\small $F_1$ (top two rows) and $\kappa$ results (bottom two rows) in three static scenes for different interpretations of data from RGB-D camera---2D keypoint detection (Zone); raw 3D point cloud (PC); raw 3D point cloud with added spheres (PC+Sph), Raw 3D point cloud with added cylinders (PC+Cyl); raw 3D point cloud with added bounding box (PC+Box).}
    \label{fig:exp1_results}
    \vspace*{-2mm}
\end{figure}

As shown in \fig{\ref{fig:exp1_results}}, the trends are very similar in both metrics. The 2D detection is outperformed by others in both metrics and in both spaces.
The results of the other interpretations indicate that the enhancement of the point cloud by keypoints improves the performance. The addition of cylinders has the highest median for both metrics and for both spaces. Moreover, for the `human' space, the addition of cylinders has the highest maxima as well. For the `robot' space, the addition of bounding boxes has the highest maxima. The addition of spheres has the worst results from the additions, but still outperforms the raw 3D point cloud.

\begin{figure}[tbp]
    \centering
    \begin{subfigure}{0.23\textwidth}
    \includegraphics[trim={2cm 2cm 2cm 6cm},clip,width=\textwidth]{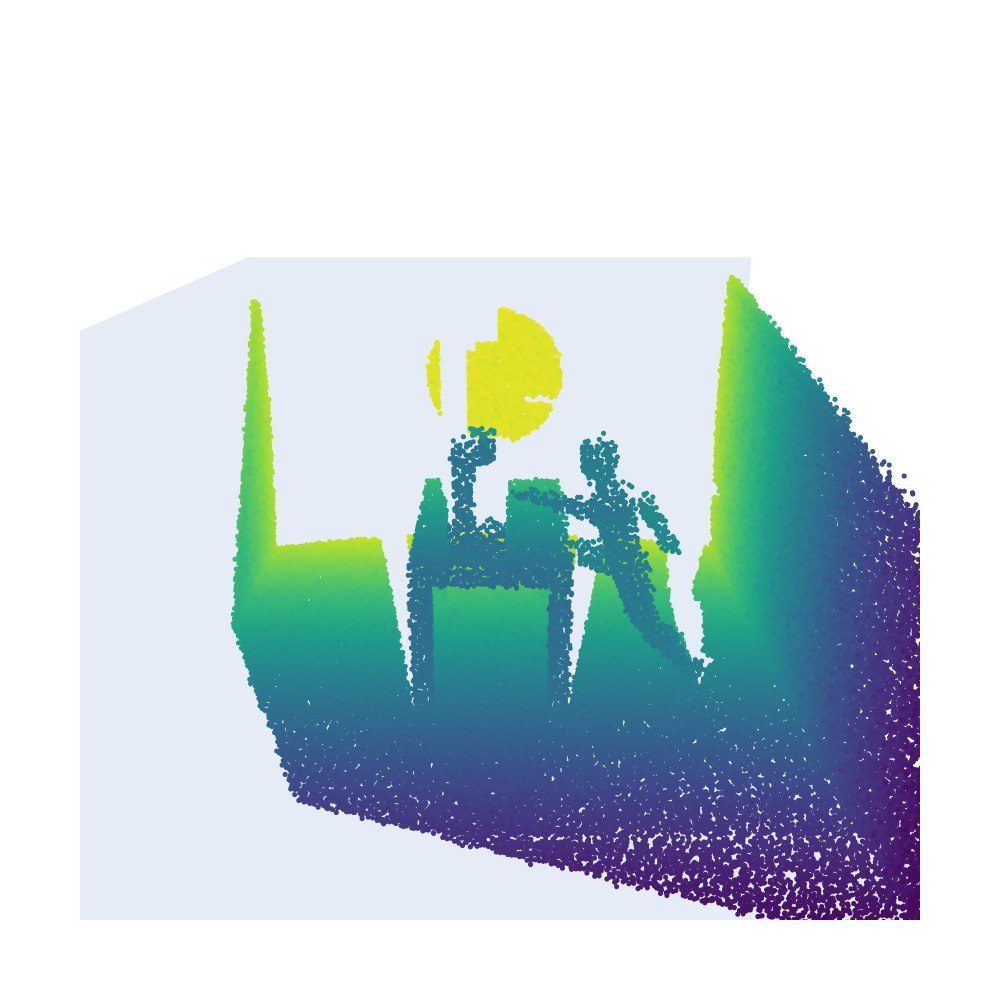}
    \caption{\footnotesize Highest $F_1$ value for Scene 1.}
    \end{subfigure}
    \begin{subfigure}{0.23\textwidth}
    \includegraphics[trim={2cm 2cm 2cm 6cm},clip,width=\textwidth]{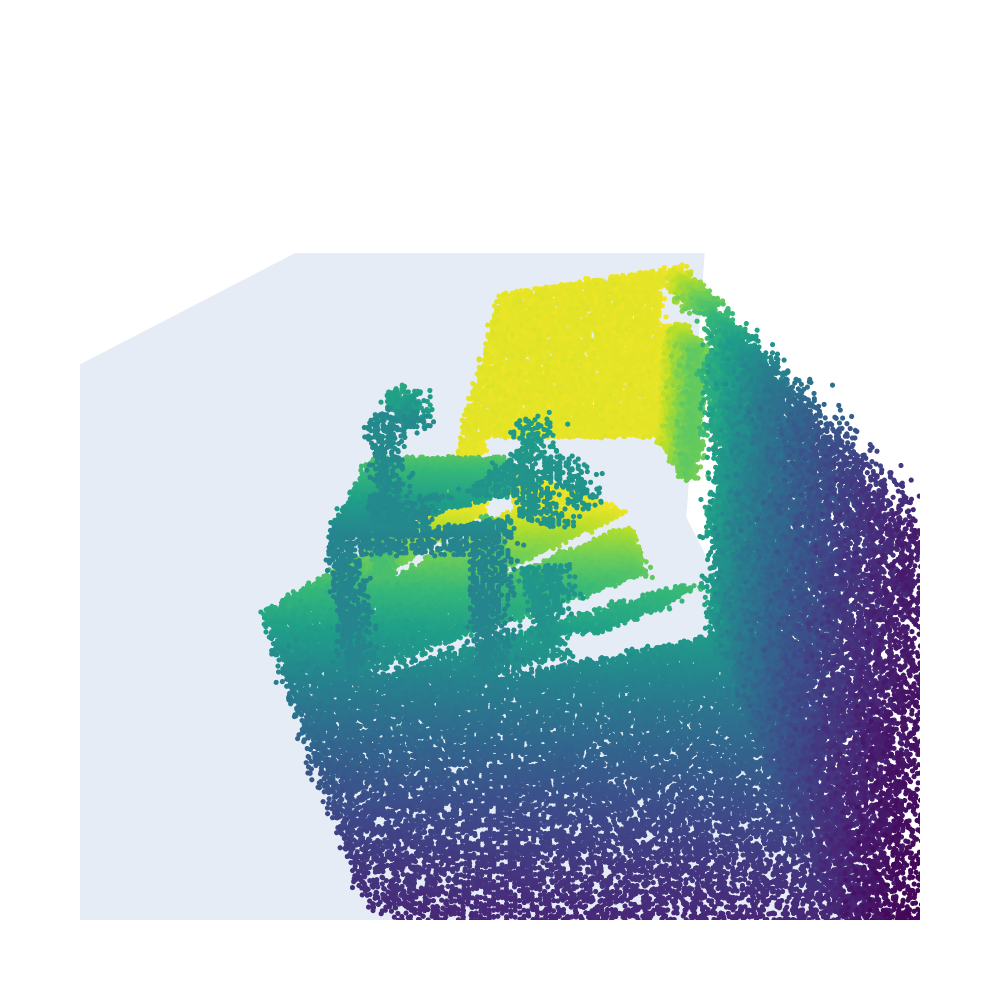}
    \caption{\footnotesize Highest $F_1$ value for Scene 2.}
    \end{subfigure}
    \begin{subfigure}{0.23\textwidth}
    \includegraphics[trim={2cm 2cm 2cm 6cm},clip,width=\textwidth]{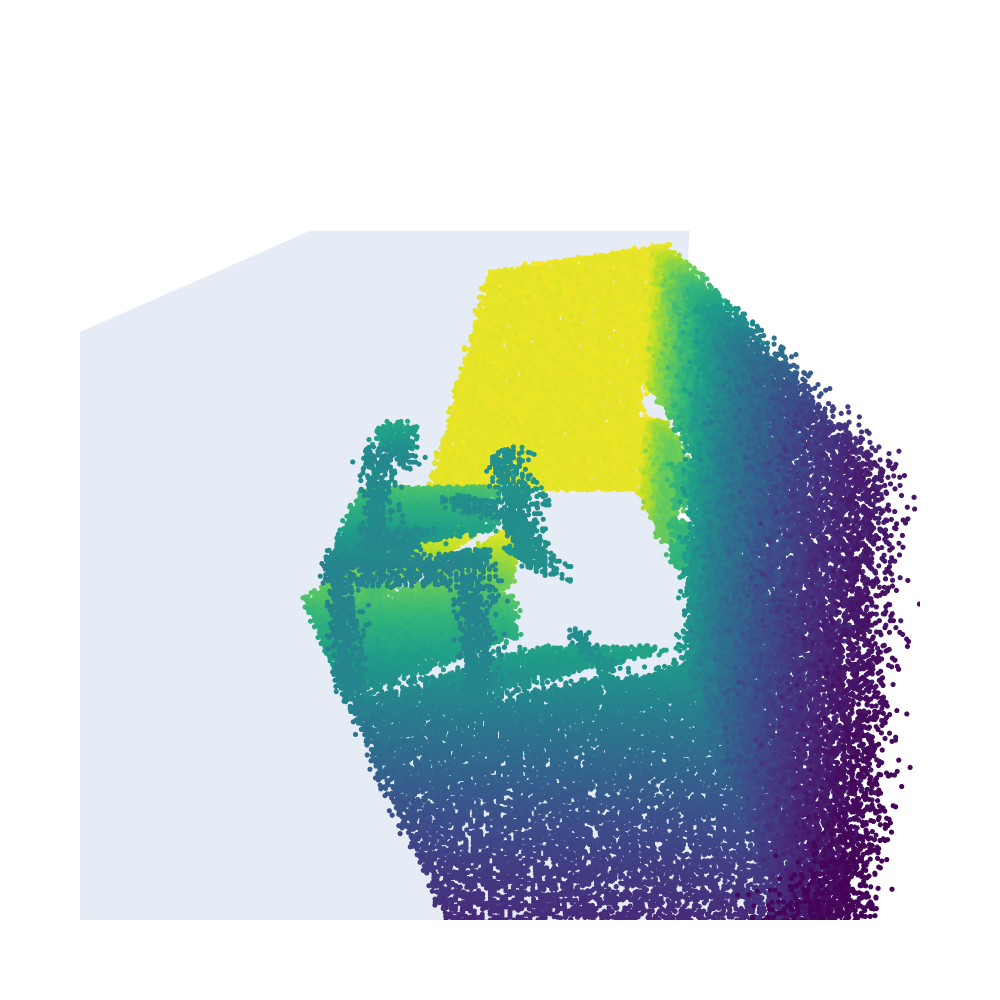}
    \caption{\footnotesize Highest $\kappa$ value for Scene 1.}
    \end{subfigure}
        \begin{subfigure}{0.23\textwidth}
    \includegraphics[trim={2cm 2cm 2cm 6cm},clip,width=\textwidth]{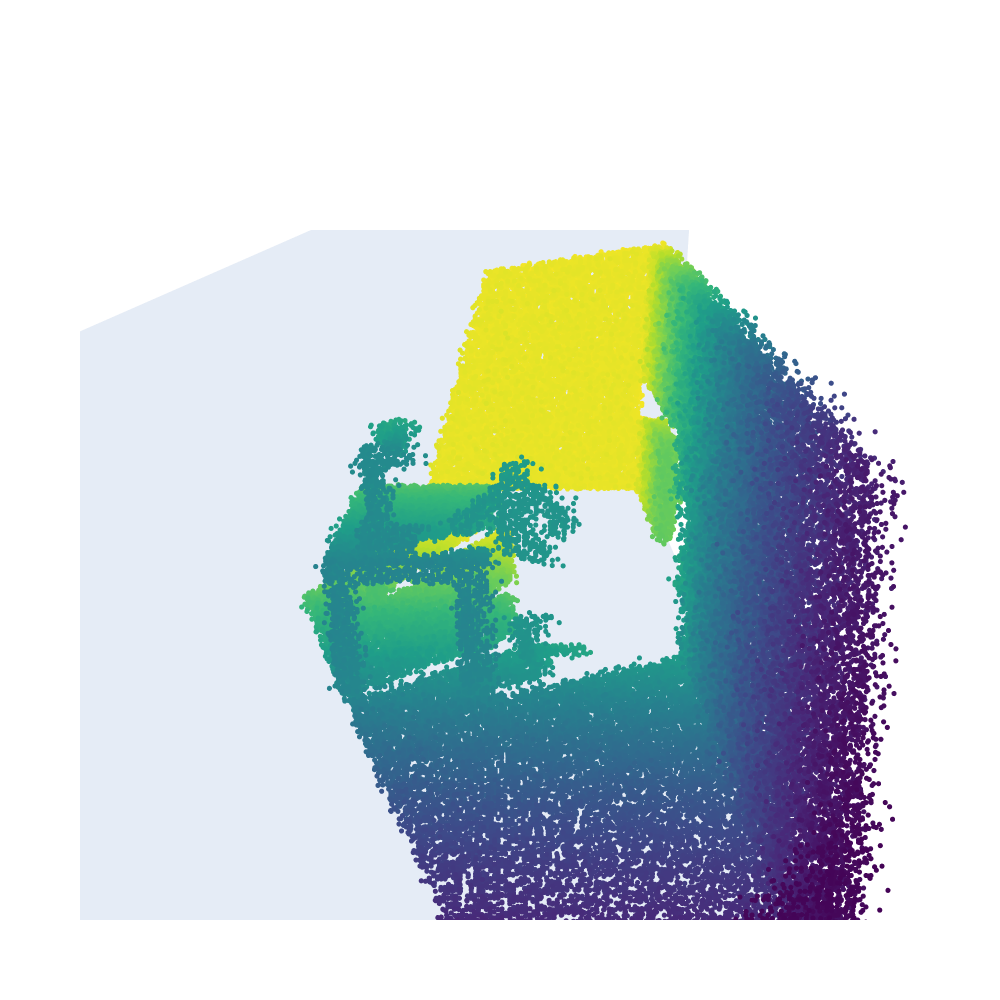}
    \caption{\footnotesize Highest $\kappa$ value for Scene 2.}
    \end{subfigure}
    \caption{\small Point clouds for the best RGB-D camera poses for `robot' space; x-axis coordinate is color-coded for better visibility.}
    \label{fig:exp1_octomap}
    \vspace*{-3mm}
\end{figure}

In addition, we can analyze the RGB-D camera placement for `robot' space coverage. Figure \ref{fig:exp1_heatmap} shows heatmaps, where each camera position is represented by its highest metric value of the five orientations for the enhanced point cloud by keypoint cylinders in Scene 3. As can be seen, there are many solid placements based on the $F_1$ metric all around the walls. Moreover, it is clearly visible that placements close to the human that obscures the operator have a much lower $F_1$ score than the other placements around. The $\kappa$ metric results emphasize the positions only on the right side of the perimeter with the best positions on the front wall. 
The difference between the metrics suggests that the positions on the right and front side detect empty voxels better than those on the left and back side.
The simulated point clouds for the RGB-D camera placement with the highest `robot' space values of $F_1$ and $\kappa$ for Scene 1 and Scene 2 are shown in \fig{\ref{fig:exp1_octomap}}. The camera is placed on the right wall (in the same notation as for the heatmap) for the highest value of $F_1$ in Scene 1. In other cases, the camera is placed on the front wall, and the placement for the highest value of $\kappa$ is the same for both scenes.  

\subsection{Multi-sensor coverage experiment (Exp 2)}
The second experiment aims to evaluate the integration of several different sensors together and to determine how the combinations improve the coverage of the space. We integrated a pressure pad, a robot proximity cover, and an RGB-D camera placed on the robot base looking front. We tested 24 pressure pad positions, three proximity cover ranges (0.1, 0.2, 0.3 m), and 9 different orientations for the RGB-D camera (from -40$^\circ$ to 40$^\circ$ around the z-axis with a step of 10$^\circ$). We evaluated the 7 possible combinations of the sensors---a triplet of sensors, 3 pairs of sensors, and 3 individual sensors. Unlike in the previous experiment, we integrated the known pose of the robot into the occupied representation to show that both possibilities are available. 

\begin{figure}[tbp]
    \centering
    \includegraphics[width=0.49\textwidth]{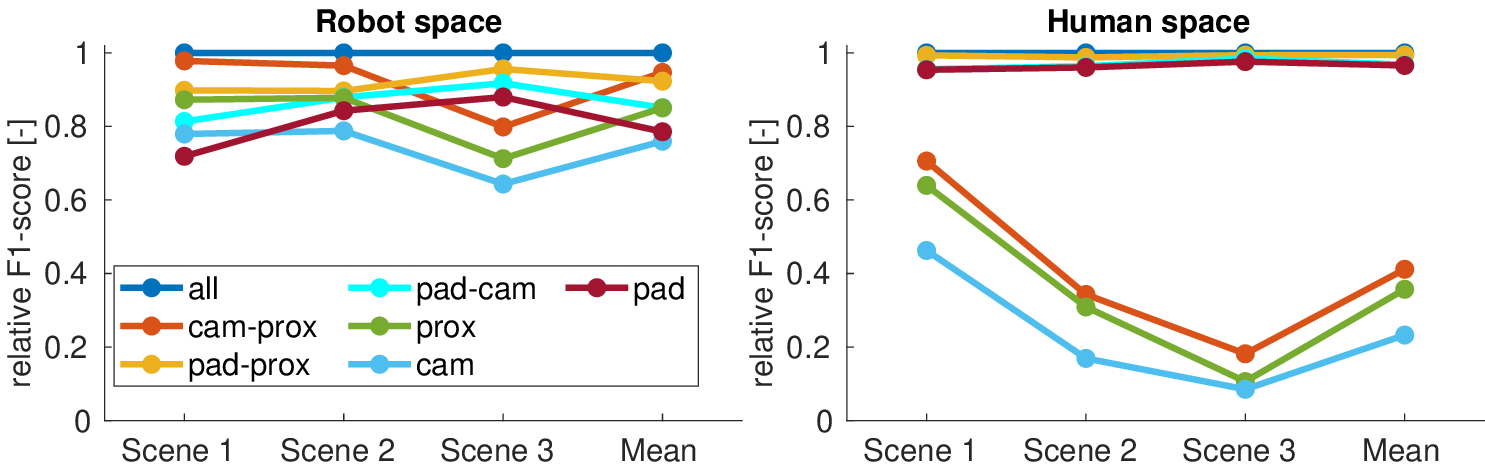}
    \includegraphics[width=0.49\textwidth]{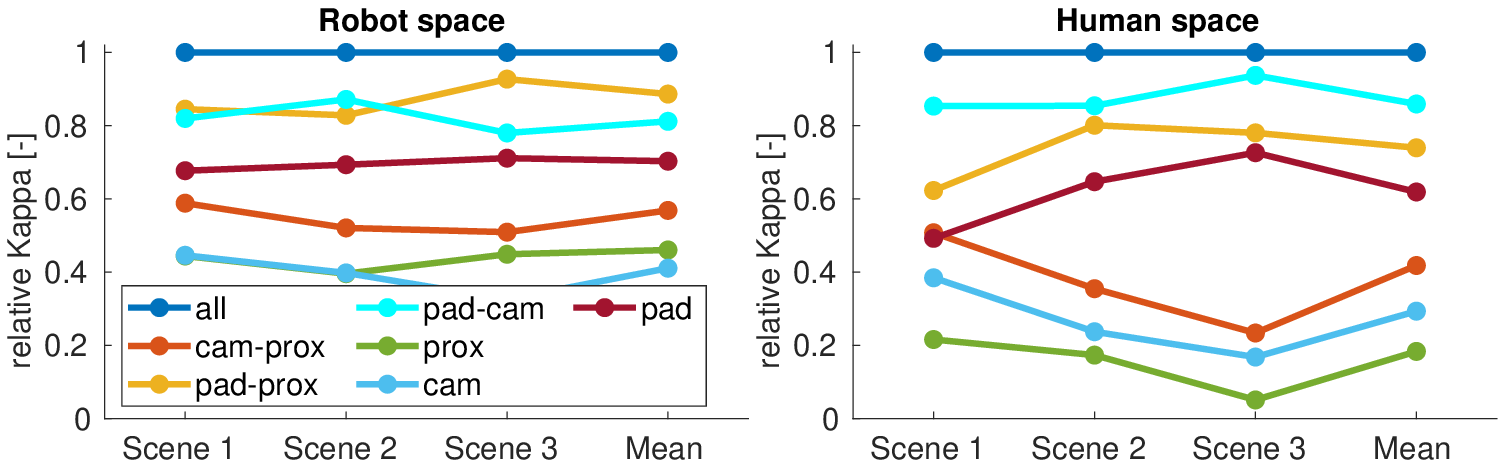}
    \caption{\small Maxima of $F_1$ (top) and Kappa (bottom) values in three static scenes and mean over scenes for all possible combinations of a pressure pad (pad), a robot proximity cover (prox), and an RGB-D camera (cam) sensors.}
    \label{fig:exp2_results}
    \vspace*{-4mm}
\end{figure}

\begin{figure}[htb]
    \centering
    \begin{subfigure}{0.23\textwidth}
    \includegraphics[trim={0 0 0 4cm},clip,width=\textwidth]{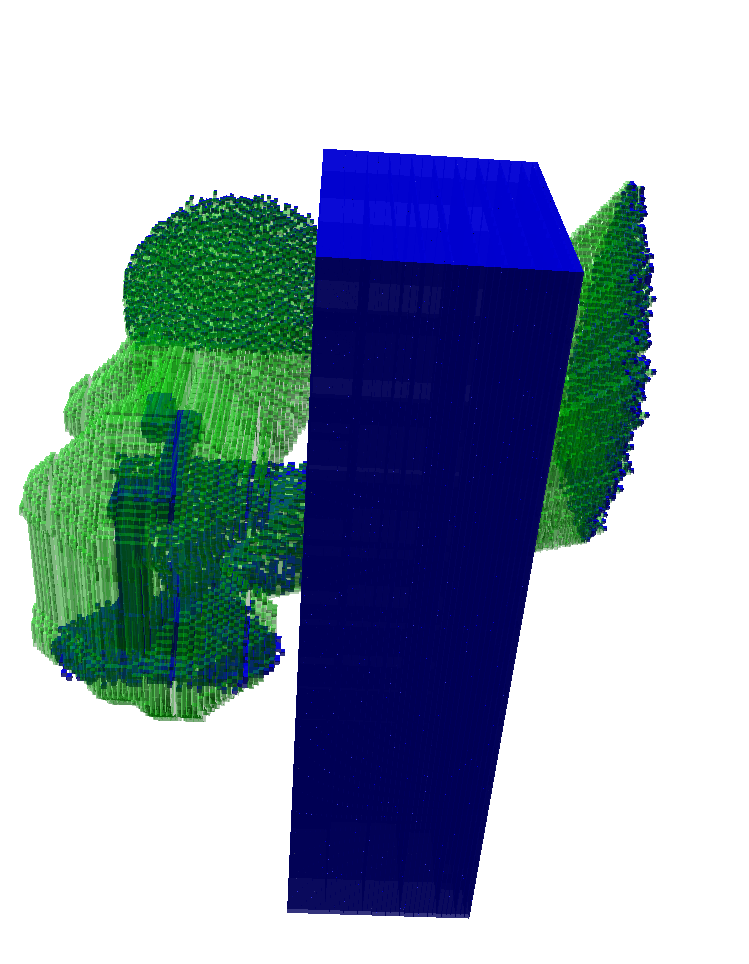}
    \begin{minipage}[t][0.8cm][t]{\linewidth}
    \vspace{-0.5cm}
    \caption{\footnotesize Highest $F_1$ and $\kappa$ for all three sensors together.}
    \end{minipage}
    \end{subfigure}
    \begin{subfigure}{0.23\textwidth}
    \includegraphics[trim={0 0 0 4cm},clip,width=\textwidth]{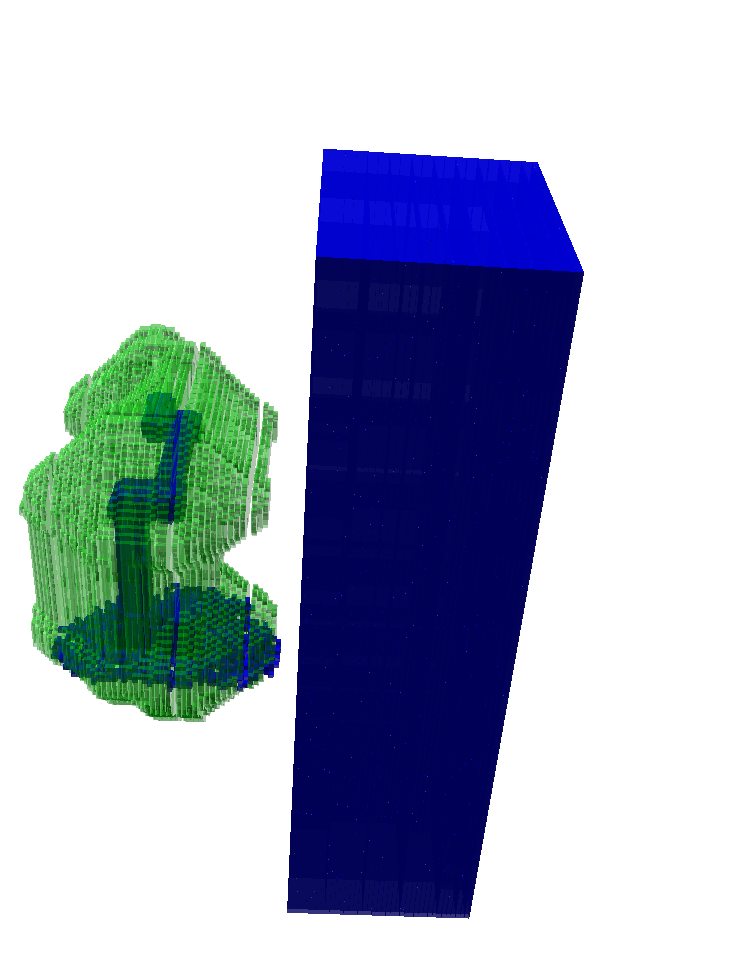}
    \begin{minipage}[t][0.8cm][t]{\linewidth}
    \vspace{-0.5cm}
    \caption{\footnotesize Highest $F_1$ and $\kappa$ for the proximity cover and the pressure pad.}
    \end{minipage}
    \end{subfigure}
    \begin{subfigure}{0.23\textwidth}
    \includegraphics[trim={0 0 0 4cm},clip,width=\textwidth]{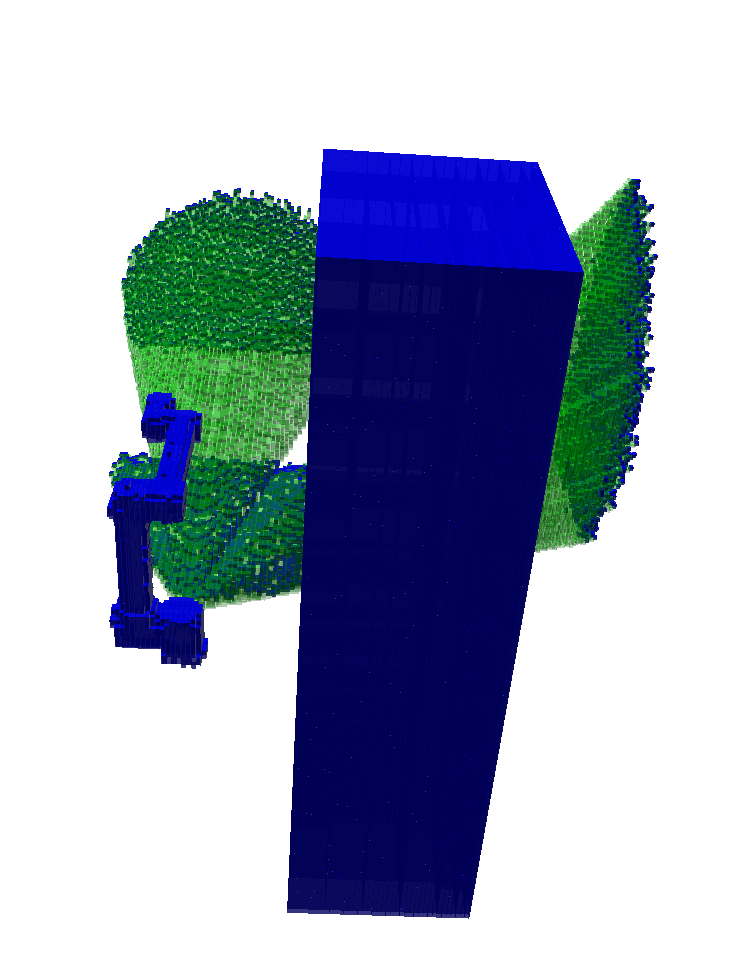}
    \caption{\footnotesize Highest $F_1$ for the pressure pad and the RGB-D camera.}
    \end{subfigure}
    \begin{subfigure}{0.23\textwidth}
    \includegraphics[trim={0 0 0 4cm},clip,width=\textwidth]{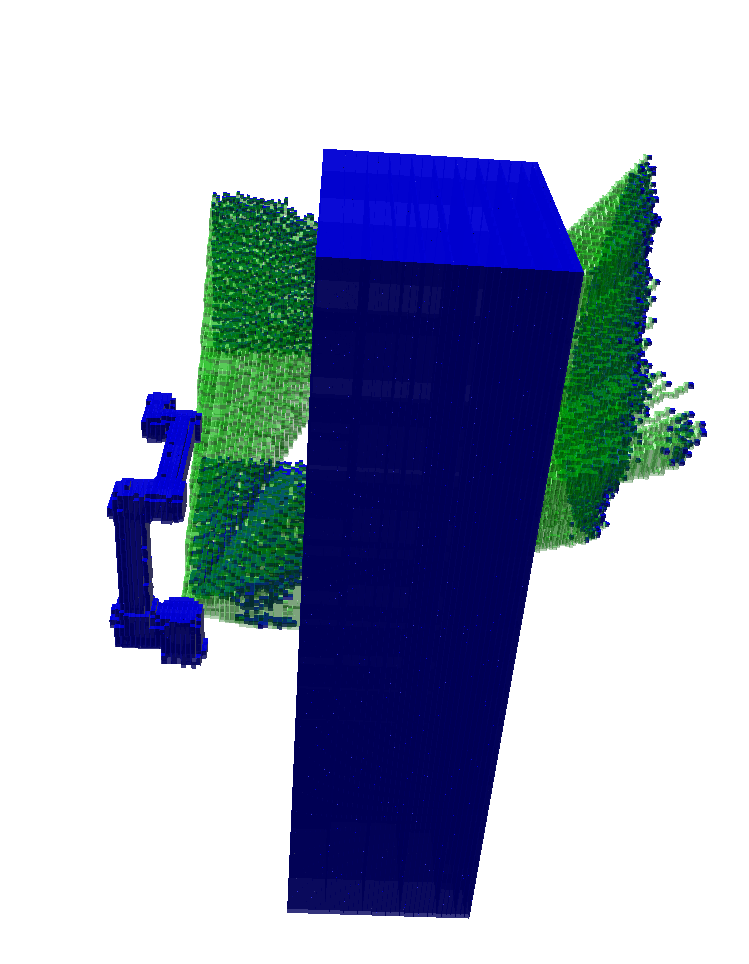}
    \caption{\footnotesize Highest $\kappa$ for the pressure pad and the RGB-D camera.}
    \end{subfigure}
    \caption{\small Occupancy voxels of occupied space (blue) and free space (green) for best variants of sensor combinations for both metrics.}
    \label{fig:exp2_octomap}
    \vspace*{-3mm}
\end{figure}
In this scenario, we compare only the maximum $F_1$ and $\kappa$ as we look for the best placement only for the combination of sensors. Figure \ref{fig:exp2_results} shows the maxima of the $F_1$ and $\kappa$ metrics for the sensor combinations relative to the triplet values.

From the F-score point of view, we can see that the combination of all three sensors does not outperform the other sensor combinations remarkably. Moreover, in the case of the evaluation of the `human' space, the $F_1$ values are almost the same for all combinations containing the pressure pad. Interestingly, the pair without the pressure pad is the second-best combination for the `robot' space, followed by the combination of the pressure pad and the robot proximity cover. The $F_1$ results show the difference between the covered spaces by individual sensors, e.g., the pressure pad is more suitable for the `human' space than the two other sensors and even their combination.

For the $\kappa$ results, the trends are very similar to those of the $F_1$ results. However, the relative differences between combinations are larger, mainly for the `robot' space. The triplet is followed by the pairs containing the pressure pad and the pressure pad alone.
These results suggest that using a pair of sensors instead of all three sensors can have a better price--performance ratio in these scenarios.

Figure \ref{fig:exp2_octomap} shows the occupancy voxels for the sensor combinations. For each combination, the variant with the highest `human' space values of $F_1$ and $\kappa$ for Scene 2 is shown.

\subsection{Dynamic scene experiment (Exp 3)}
In the third experiment, we look for the best position of the LIDAR sensor in the dynamic scene represented by 27 snapshots of the scene where the human and the robot were moving. We computed the coverage of the LIDAR sensor in the same 172 positions as those used in Exp 1.

\begin{figure}[htbp]
    \centering
    \includegraphics[trim={7cm 3cm 0 3cm},clip,width=0.48\textwidth]{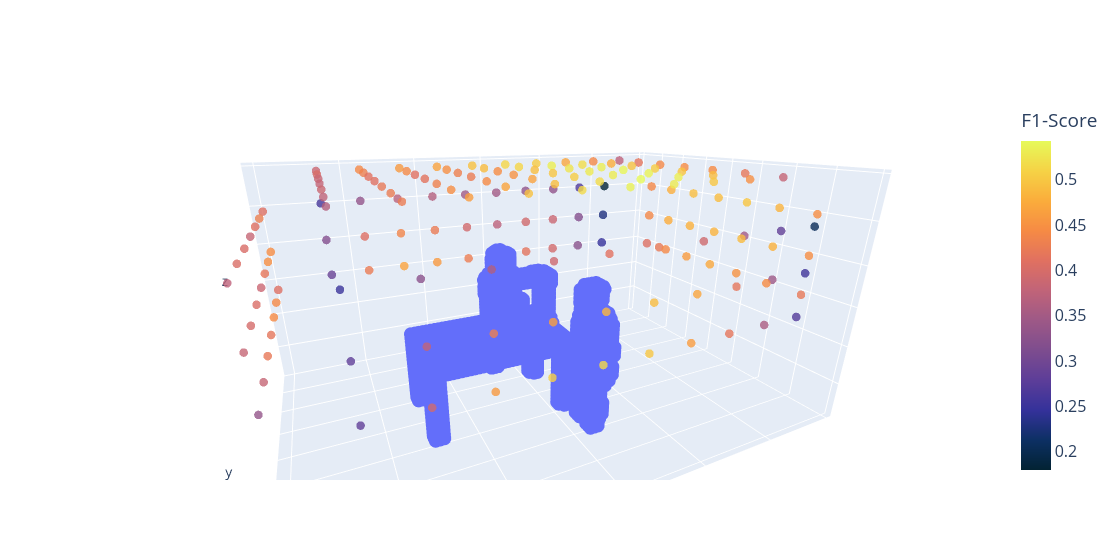}
    \includegraphics[trim={7cm 3cm 0 3cm},clip,width=0.48\textwidth]{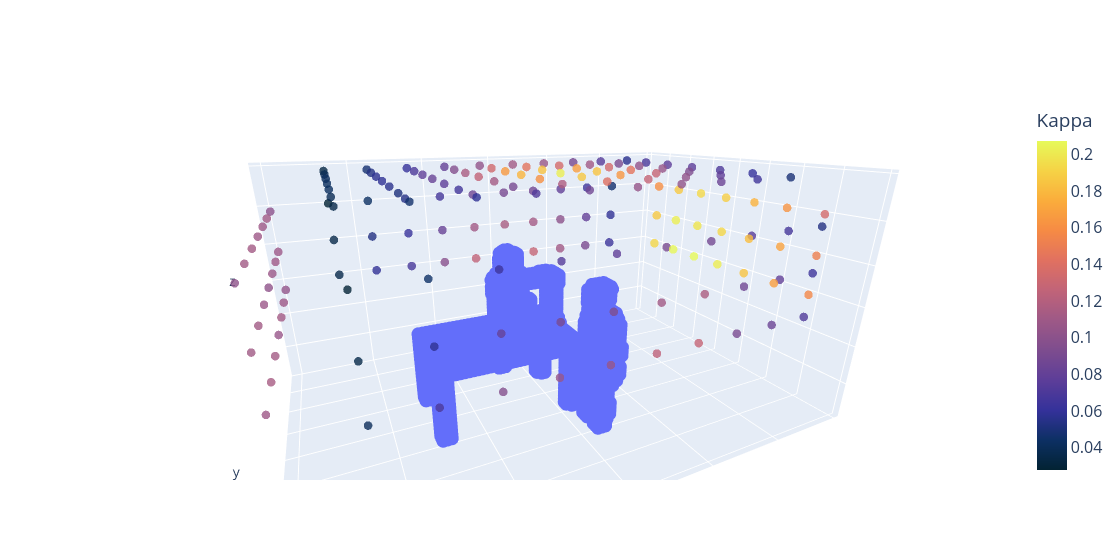}
    \caption{\small LIDAR sensor placement heatmap for `robot' space in dynamic scene; $F_1$ (top) and $\kappa$ (bottom) values.}
    \label{fig:exp3_heatmap}
    \vspace*{-2mm}
\end{figure}

\begin{figure}[htbp]
    \centering
        \begin{subfigure}{0.23\textwidth}
    \includegraphics[trim={2cm 2cm 2cm 6cm},clip,width=\textwidth]{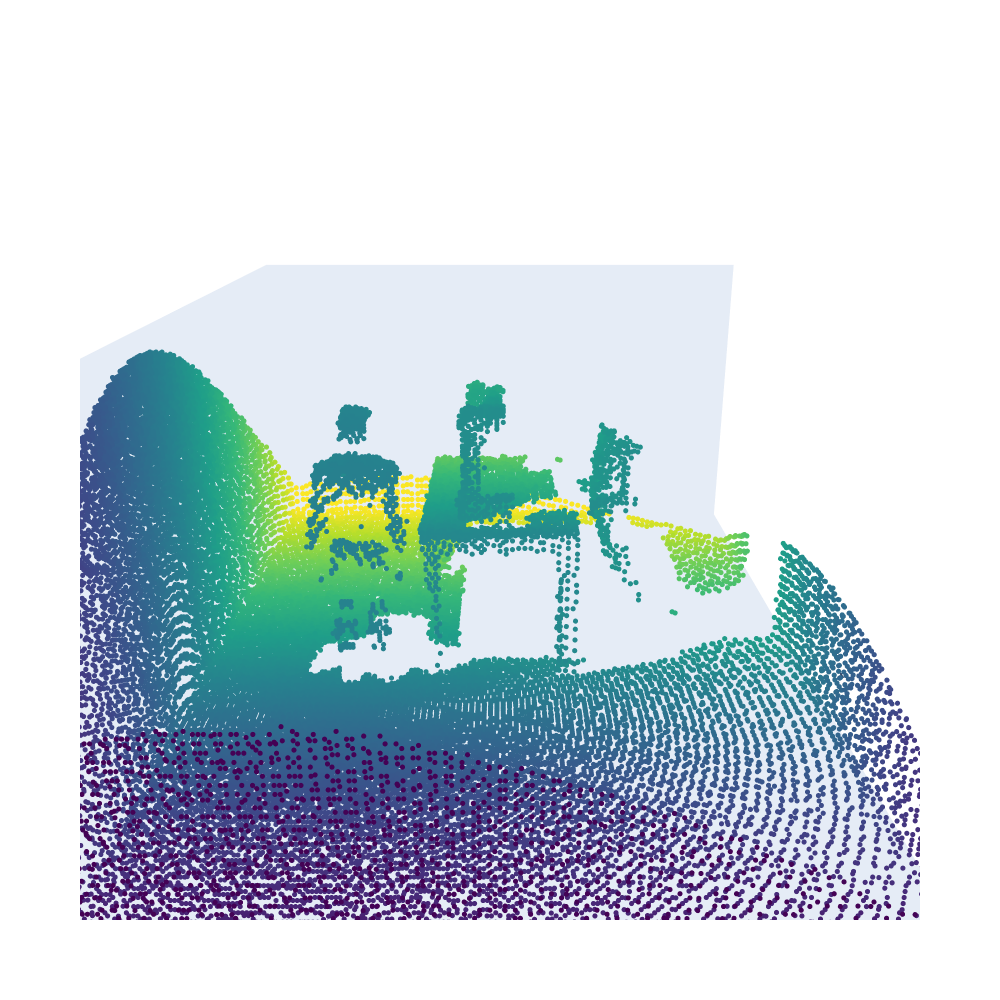}
    \caption{\footnotesize Snapshot 1 for the highest $F_1$.}
     \end{subfigure}
             \begin{subfigure}{0.23\textwidth}
    \includegraphics[trim={2cm 2cm 2cm 6cm},clip,width=\textwidth]{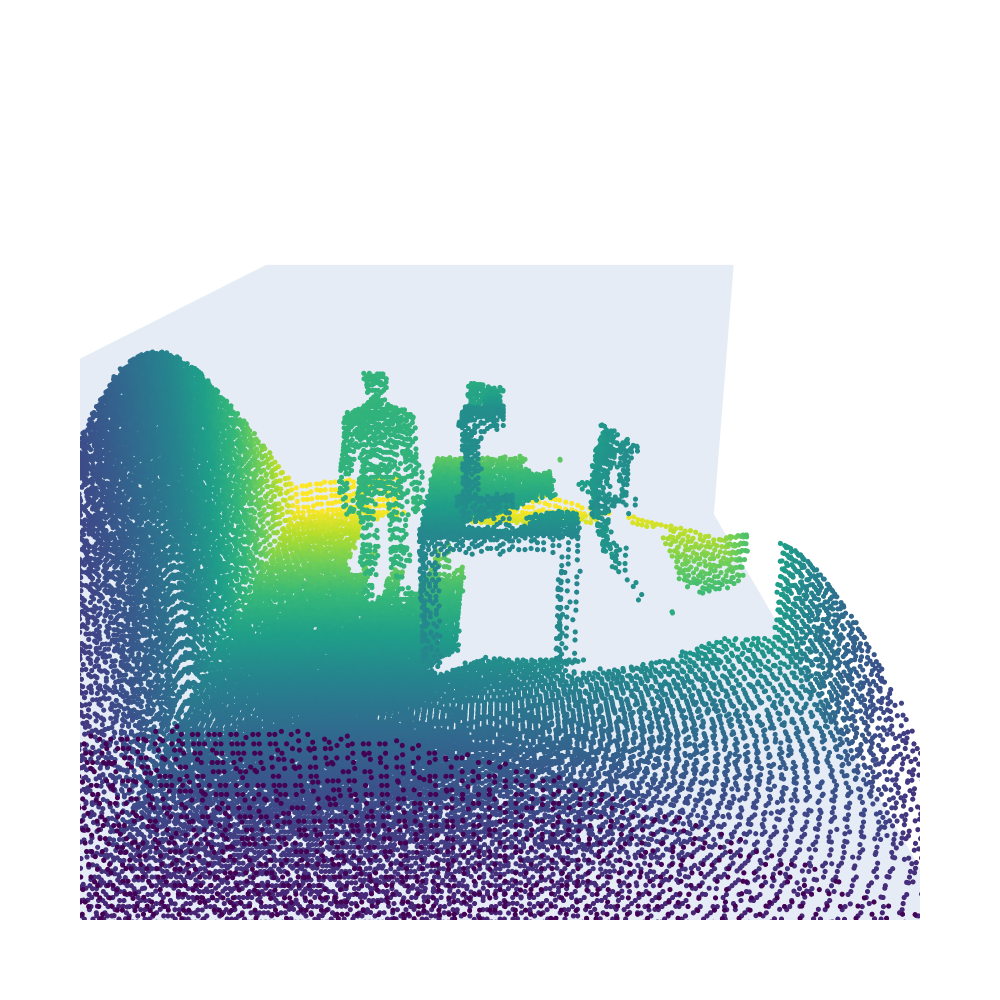}
        \caption{\footnotesize Snapshot 2 for the highest $F_1$.}
    \end{subfigure}
    \begin{subfigure}{0.23\textwidth}
    \includegraphics[trim={2cm 2cm 2cm 6cm},clip,width=\textwidth]{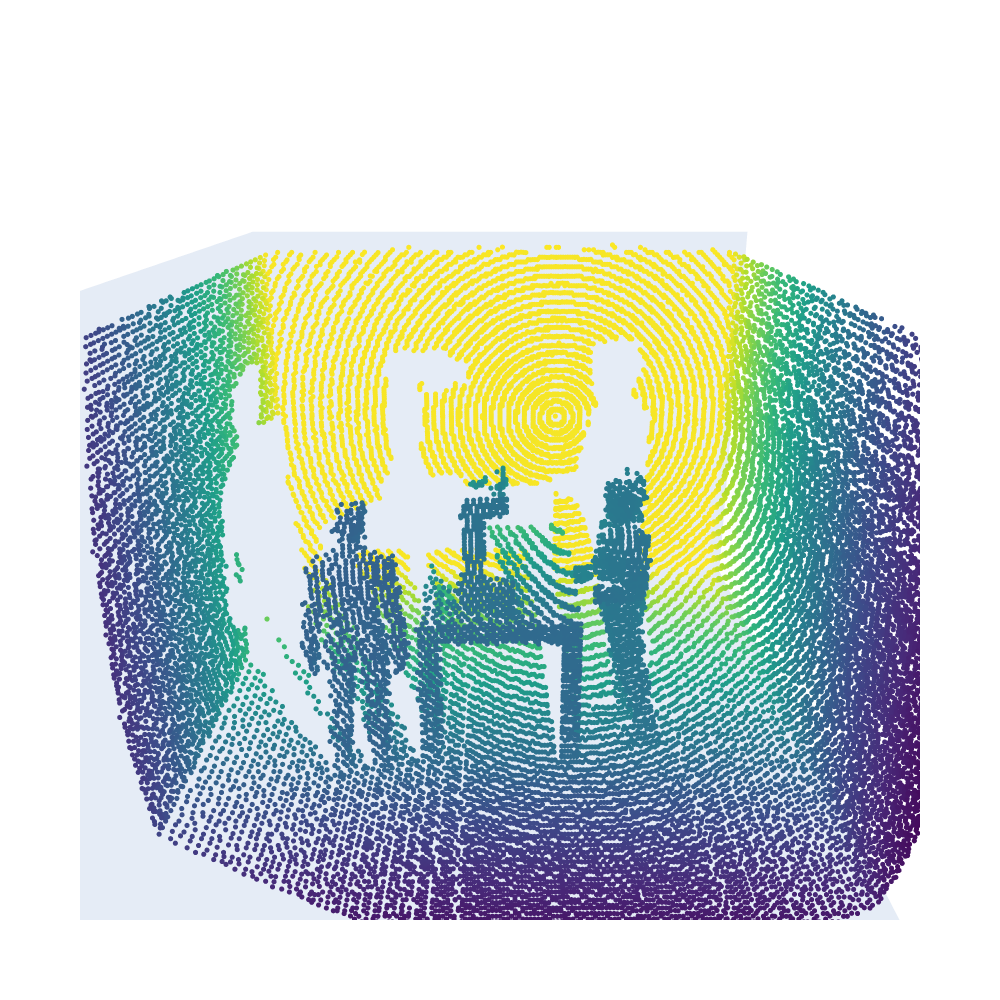}
        \caption{\footnotesize Snapshot 1 for the highest $\kappa$.}
    \end{subfigure}
    \begin{subfigure}{0.23\textwidth}
    \includegraphics[trim={2cm 2cm 2cm 6cm},clip,width=\textwidth]{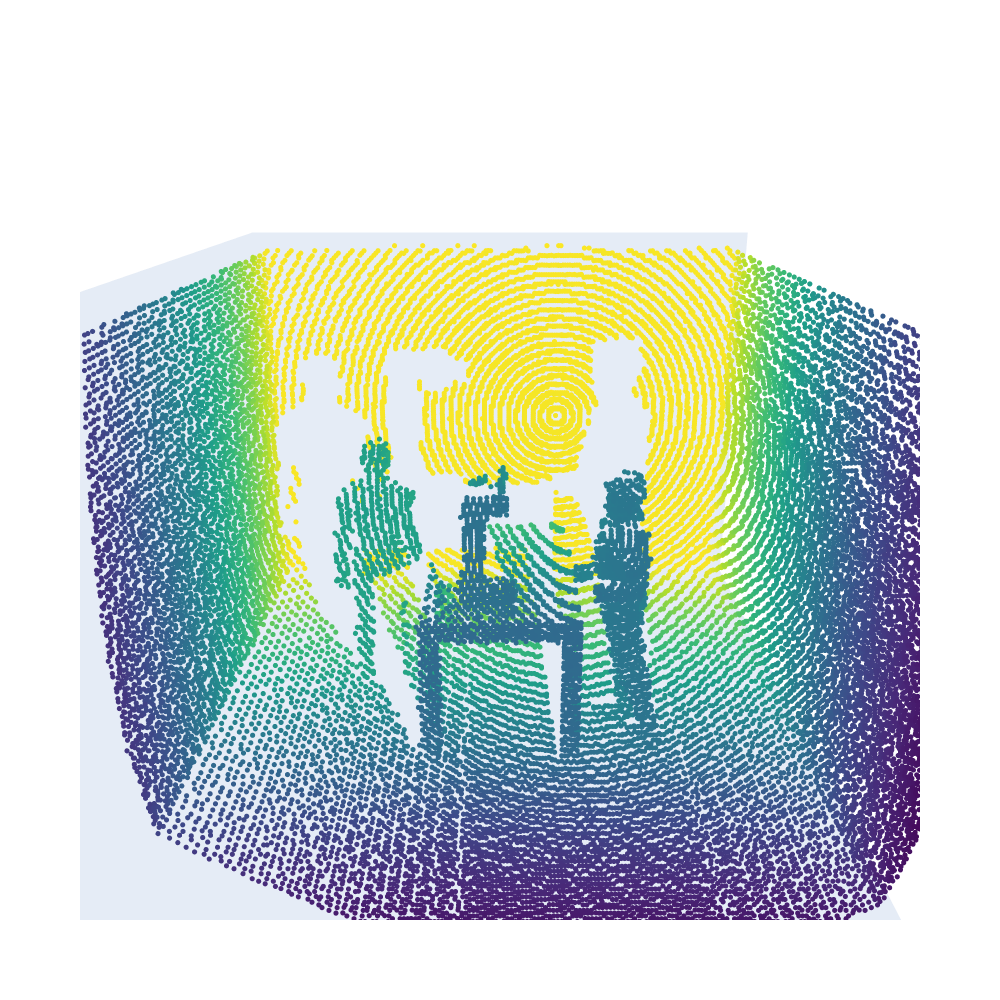}
        \caption{\footnotesize Snapshot 2 for the highest $\kappa$.}
        \end{subfigure}
    \caption{\small Point clouds for the LIDAR sensor position with the highest `robot' space values of $F_1$ (sensor on the ceiling) and $\kappa$ (sensor on the wall) for the dynamic scene; x-axis coordinate is color-coded for better readability.}
    \label{fig:exp3_octomap}
    \vspace*{-2mm}
\end{figure}

In this experiment, we evaluate only the LIDAR placement for the `robot' space coverage. Similarly to the first experiment, Figure \ref{fig:exp3_heatmap} shows heatmaps, where each LIDAR position is represented by its highest metric value. As can be seen, there are differences between the best positions based on the $F_1$ and $\kappa$ values. The $F_1$ results suggest placing the LIDAR in the right half of the ceiling. From the ceiling positions, the $\kappa$ results emphasize only a few places close to the center of the ceiling, and also propose placements on the right wall.
The simulated point clouds for the LIDAR placement with the highest `robot' space values of $F_1$ and $\kappa$ are shown in \fig{\ref{fig:exp3_octomap}} for two different snapshots. As can be seen, the best position based on the $F_1$ metric is on the ceiling. On the other hand, the position with the highest $\kappa$ values is on the right wall (in the same notation as for the heatmap).

\section{Conclusion}
We introduced the definition of the perirobot space (PeRS)---the monitored region of interest for human-robot interaction where sensor data are represented as occupancy information. 
This study presented the formalization and evaluation of PeRS. 
We demonstrated its use in an RGB-D camera placement experiment, a multi-sensor coverage experiment, and a dynamic scene experiment. 
The occupancy representation allowed us to compare the effectiveness of various sensor setups and we used the well-established metrics of $F_1$ and $\kappa$ score to evaluate the coverage of regions of interest.
Therefore, our approach can serve as a prototyping tool to establish the sensor setup that provides the most efficient coverage with respect to the given metrics and sensor representations. 
For that reason, we made the implementation publicly available\footnote{\url{https://github.com/ctu-vras/perirobot-space}}. 
\section{Discussion}
The central idea of our approach is the simple evaluation of integrated sensors with different properties.
We consider this work a first step that could be extended in multiple directions. 
Our approach can be interpreted in two ways. 
First, given a task, determine the optimal sensor setup. 
Second, given a sensor setup, what is the best setup of the task---e.g. the position of the robot and operator in the workspace.
For our presented multi-sensor grid search to determine the optimal sensor setup, we could consider additional criteria (e.g., space limitations or sensor costs to find the best price--performance ratio).

Our approach can capture the differences between sensors of the same type, as revealed by different suggestions for placing the LIDAR sensor and the RGB-D camera (see Figs. \ref{fig:exp1_heatmap}, \ref{fig:exp3_heatmap}).
Moreover, the known robot pose can be integrated into the occupied space, as shown in the multi-sensor experiment (see \fig{\ref{fig:exp2_octomap}}), to disfavor sensor variants observing only the robot and not the space around.
We found a difference between the $F_1$ and $\kappa$ scores for the best placements for the sensors. This is probably caused by the $F_1$ score overlooking the correctly detected free space.

While we presented only a few sensors in this paper, 
our occupancy-based approach allows the easy addition of new sensors.
For a detailed evaluation of a sensor setup, all appropriate representations should be considered.

The presented approach dealt with regions of interest surrounding the robot and the human. 
However, the instantaneous robot speed could be taken into account to determine an appropriate surrounding volume for the robot to be considered as the monitored space. 

Other possible additions include additional sensors (e.g., moving sensors, sensors on humans), further representations (e.g., swept volumes for the human), taking into account the human attention (e.g., by tracking human gaze and incorporating this attention as a factor), using the occupancy representation of sensors for control (e.g., evasion of the occupied space).
Also, we would like to extend the framework with a user interface to allow its easier use and optimize the code for faster computation.
Finally, we want to focus on more efficient evaluation of dynamic scenes.




\bibliographystyle{IEEEtran}
\bibliography{pers}

\begin{thebibliography}{10}
\providecommand{\url}[1]{#1}
\csname url@samestyle\endcsname
\providecommand{\newblock}{\relax}
\providecommand{\bibinfo}[2]{#2}
\providecommand{\BIBentrySTDinterwordspacing}{\spaceskip=0pt\relax}
\providecommand{\BIBentryALTinterwordstretchfactor}{4}
\providecommand{\BIBentryALTinterwordspacing}{\spaceskip=\fontdimen2\font plus
\BIBentryALTinterwordstretchfactor\fontdimen3\font minus
  \fontdimen4\font\relax}
\providecommand{\BIBforeignlanguage}[2]{{%
\expandafter\ifx\csname l@#1\endcsname\relax
\typeout{** WARNING: IEEEtran.bst: No hyphenation pattern has been}%
\typeout{** loaded for the language `#1'. Using the pattern for}%
\typeout{** the default language instead.}%
\else
\language=\csname l@#1\endcsname
\fi
#2}}
\providecommand{\BIBdecl}{\relax}
\BIBdecl

\bibitem{lasota2017survey}
P.~A. Lasota, T.~Fong, and J.~A. Shah, ``A survey of methods for safe
  human-robot interaction,'' \emph{Foundations and Trends in Robotics}, vol.~5,
  no.~4, pp. 261--349, 2017.

\bibitem{ISO10218}
{ISO, IEC}, ``{ISO 10218 Robots and robotic devices -- Safety requirements for
  industrial robots},'' {International Organization for Standardization},
  Geneva, CH, Standard, 2011.

\bibitem{ISO/TS15066}
------, ``{ISO/TS 15066 Robots and robotic devices -- Collaborative robots},''
  {International Organization for Standardization}, Geneva, CH, Standard, 2016.

\bibitem{wang2011coverage}
B.~Wang, ``Coverage problems in sensor networks: A survey,'' \emph{ACM
  Computing Surveys (CSUR)}, vol.~43, no.~4, pp. 1--53, 2011.

\bibitem{elhabyan2019coverage}
R.~Elhabyan, W.~Shi, and M.~St-Hilaire, ``Coverage protocols for wireless
  sensor networks: Review and future directions,'' \emph{Journal of
  Communications and Networks}, vol.~21, no.~1, pp. 45--60, 2019.

\bibitem{Oscadal2023}
P.~Oščádal, T.~Kot, T.~Spurný, J.~Suder, M.~Vocetka, L.~Dobeš, and
  Z.~Bobovský, ``Camera arrangement optimization for workspace monitoring in
  human--robot collaboration,'' \emph{Sensors}, vol.~23, no.~1, p. 295, 2023.

\bibitem{he2022visibility}
K.~He, R.~Newbury, T.~Tran, J.~Haviland, B.~Burgess-Limerick, D.~Kulić,
  P.~Corke, and A.~Cosgun, ``{Visibility Maximization Controller for Robotic
  Manipulation},'' \emph{IEEE Robotics and Automation Letters}, vol.~7, no.~3,
  pp. 8479--8486, 2022.

\bibitem{ibrahim2022whole}
I.~Ibrahim, F.~Farshidian, J.~Preisig, P.~Franklin, P.~Rocco, and M.~Hutter,
  ``{Whole-Body MPC and Dynamic Occlusion Avoidance: A Maximum Likelihood
  Visibility Approach},'' in \emph{2022 International Conference on Robotics
  and Automation (ICRA)}, 2022, pp. 221--227.

\bibitem{finean2021should}
M.~N. Finean, W.~Merkt, and I.~Havoutis, ``{Where Should I Look? Optimized Gaze
  Control for Whole-Body Collision Avoidance in Dynamic Environments},''
  \emph{{IEEE Robotics and Automation Letters}}, vol.~7, no.~2, pp. 1095--1102,
  2021.

\bibitem{alatise2020review}
M.~B. Alatise and G.~P. Hancke, ``A review on challenges of autonomous mobile
  robot and sensor fusion methods,'' \emph{IEEE Access}, vol.~8, pp.
  39\,830--39\,846, 2020.

\bibitem{Magnanimo2016}
V.~Magnanimo, S.~Walther, L.~Tecchia, C.~Natale, and T.~Guhl, ``{Safeguarding a
  mobile manipulator using dynamic safety fields},'' in \emph{{Intelligent
  Robots and Systems (IROS), 2016 IEEE/RSJ International Conference on}}, 2016,
  pp. {2972--2977}.

\bibitem{rashid2020local}
A.~Rashid, K.~Peesapati, M.~Bdiwi, S.~Krusche, W.~Hardt, and M.~Putz, ``{Local
  and global sensors for collision avoidance},'' in \emph{{2020 IEEE
  International Conference on Multisensor Fusion and Integration for
  Intelligent Systems (MFI)}}, 2020, pp. {354--359}.

\bibitem{suvsanj2020effective}
D.~Su{\v{s}}anj, D.~Pin{\v{c}}i{\'c}, and K.~Lenac, ``Effective area coverage
  of 2d and 3d environments with directional and isotropic sensors,''
  \emph{IEEE Access}, vol.~8, pp. 185\,595--185\,608, 2020.

\bibitem{Marvel2013}
J.~A. Marvel, ``{Performance metrics of speed and separation monitoring in
  shared workspaces},'' \emph{{IEEE Transactions on Automation Science and
  Engineering}}, vol.~{10}, no.~2, pp. {405--414}, 2013.

\bibitem{marvel2017implementing}
J.~A. Marvel and R.~Norcross, ``{Implementing speed and separation monitoring
  in collaborative robot workcells},'' \emph{{Robotics and Computer-Integrated
  Manufacturing}}, vol.~{44}, pp. {144--155}, 2017.

\bibitem{fabrizio2016real}
F.~Flacco and A.~De~Luca, ``{Real-time computation of distance to dynamic
  obstacles with multiple depth sensors},'' \emph{{IEEE Robotics and Automation
  Letters}}, vol.~{2}, no.~1, pp. {56--63}, 2016.

\bibitem{Lacevic2010}
B.~Lacevic and P.~Rocco, ``{Kinetostatic danger field-a novel safety assessment
  for human-robot interaction},'' in \emph{{Intelligent Robots and Systems
  (IROS), 2010 IEEE/RSJ International Conference on}}, 2010, pp. {2169--2174}.

\bibitem{Zanchettin2016}
A.~M. Zanchettin, N.~M. Ceriani, P.~Rocco, H.~Ding, and B.~Matthias, ``{Safety
  in human-robot collaborative manufacturing environments: Metrics and
  control},'' \emph{{IEEE Transactions on Automation Science and Engineering}},
  vol.~{13}, no.~2, pp. {882--893}, 2016.

\bibitem{Polverini2017}
M.~P. Polverini, A.~M. Zanchettin, and P.~Rocco, ``{A computationally efficient
  safety assessment for collaborative robotics applications},'' \emph{{Robotics
  and Computer-Integrated Manufacturing}}, vol.~{46}, pp. {25--37}, 2017.

\bibitem{byner2019dynamic}
C.~Byner, B.~Matthias, and H.~Ding, ``{Dynamic speed and separation monitoring
  for collaborative robot applications--concepts and performance},''
  \emph{{Robotics and Computer-Integrated Manufacturing}}, vol.~{58}, pp.
  {239--252}, 2019.

\bibitem{scalera2021optimal}
L.~Scalera, R.~Vidoni, and A.~Giusti, ``{Optimal scaling of dynamic safety
  zones for collaborative robotics},'' in \emph{{2021 IEEE International
  Conference on Robotics and Automation (ICRA)}}, 2021, pp. {3822--3828}.

\bibitem{halme2018review}
R.-J. Halme, M.~Lanz, J.~K{\"a}m{\"a}r{\"a}inen, R.~Pieters, J.~Latokartano,
  and A.~Hietanen, ``{Review of vision-based safety systems for human-robot
  collaboration},'' \emph{{Procedia CIRP}}, vol.~{72}, pp. {111--116}, 2018.

\bibitem{kumar2019speed}
S.~Kumar, S.~Arora, and F.~Sahin, ``{Speed and separation monitoring using
  on-robot time-of-flight laser-ranging sensor arrays},'' in \emph{{2019 IEEE
  15th International Conference on Automation Science and Engineering (CASE)}},
  2019, pp. {1684--1691}.

\bibitem{qi2022safe}
K.~Qi, Z.~Song, and J.~S. Dai, ``{Safe physical human-robot interaction: A
  quasi whole-body sensing method based on novel laser-ranging sensor ring
  pairs},'' \emph{{Robotics and Computer-Integrated Manufacturing}}, vol.~{75},
  p. {102280}, 2022.

\bibitem{Navarro2021}
S.~E. Navarro, S.~M{\"u}hlbacher-Karrer, H.~Alagi, H.~Zangl, K.~Koyama,
  B.~Hein, C.~Duriez, and J.~R. Smith, ``Proximity perception in human-centered
  robotics: A survey on sensing systems and applications,'' \emph{IEEE
  Transactions on Robotics}, vol.~38, no.~3, pp. 1599--1620, 2021.

\bibitem{Roncone2016}
A.~Roncone, M.~Hoffmann, U.~Pattacini, L.~Fadiga, and G.~Metta, ``Peripersonal
  space and margin of safety around the body: learning tactile-visual
  associations in a humanoid robot with artificial skin,'' \emph{{PLoS} {ONE}},
  vol.~11, no.~10, pp. 1--32, 2016.

\bibitem{siciliano2008springer}
B.~Siciliano, O.~Khatib, and T.~Kr{\"o}ger, \emph{Springer handbook of
  robotics}.\hskip 1em plus 0.5em minus 0.4em\relax Springer, 2008, vol. 200.

\bibitem{Haddadin2017}
S.~Haddadin, A.~De~Luca, and A.~Albu-Sch{\"a}ffer, ``{Robot collisions: A
  survey on detection, isolation, and identification},'' \emph{{IEEE
  Transactions on Robotics (T-RO}}, vol.~33, no.~6, pp. 1292--1312, 2017.

\bibitem{hornung2013octomap}
A.~Hornung, K.~M. Wurm, M.~Bennewitz, C.~Stachniss, and W.~Burgard, ``{OctoMap:
  An efficient probabilistic 3D mapping framework based on octrees},''
  \emph{{Autonomous robots}}, vol.~{34}, no.~3, pp. {189--206}, 2013.

\end{thebibliography}

\end{document}